\newif\ifarxiv
\arxivtrue

\RequirePackage{fix-cm}

\documentclass[twocolumn]{svjour3}          

\smartqed  

\usepackage{natbib}         
\usepackage{microtype}      
\usepackage{cite}
\usepackage{paralist}		
\usepackage{color}		
\usepackage{times}
\usepackage{amsfonts}		
\usepackage{amsmath}		
\usepackage{amssymb}
\usepackage{latexsym}
\usepackage{graphicx}		
\usepackage{placeins}
\usepackage{times}
\usepackage{epsfig}
\usepackage[dvipsnames]{xcolor}
\usepackage{tabularx}
\usepackage{algorithm}
\usepackage{algorithmicx}
\usepackage{algpseudocode}
\usepackage{enumitem}
\usepackage[export]{adjustbox}
\usepackage{arydshln} 
\usepackage{color,soul}
\usepackage[
    pagebackref=true,
    breaklinks=true,
    bookmarks=false,
    colorlinks,
    linkcolor=blue,
    urlcolor=blue,
    citecolor=blue,
    anchorcolor=blue]{hyperref}
\usepackage{xurl}
\usepackage{hyphenat} 
\usepackage{tablefootnote}

\makeatletter
\DeclareRobustCommand\onedot{\futurelet\@let@token\@onedot}
\def\@onedot{\ifx\@let@token.\else.\null\fi\xspace}

\makeatother

\DeclareMathOperator*{\argmin}{arg\,min}
\DeclareMathOperator*{\argmax}{arg\,max}

\newcommand*{\blarrow}{\rotatebox[origin=c]{270}{$\Rsh$}}
\newcommand{\spm}[1]{\scriptsize{$\pm$#1}}
\newcommand*{\cm}{\checkmark}
\newcommand{\f}[1]{\textbf{#1}}
\newcommand{\s}[1]{\underline{#1}}

\journalname{International Journal of Computer Vision}

\begin{document}

\title{Improving Semi-Supervised and Domain-Adaptive Semantic Segmentation with Self-Supervised Depth Estimation}


\author{Lukas Hoyer         \and
        Dengxin Dai         \and 
        Qin Wang            \and
        Yuhua Chen          \and
        Luc Van Gool
}


\institute{Lukas Hoyer \at
              ETH Zurich, Switzerland \\
              \email{lhoyer@vision.ee.ethz.ch}           
           \and
               Dengxin Dai \at
              ETH Zurich, Switzerland \& MPI for Informatics, Germany\\
              \email{dai@vision.ee.ethz.ch} \and 
           Qin Wang \at
              ETH Zurich, Switzerland \\
              \email{qwang@ethz.ch}
           \and
           Yuhua Chen \at
              ETH Zurich, Switzerland \\
              \email{yuhua.chen@vision.ee.ethz.ch}
           \and
           Luc Van Gool \at
              ETH Zurich, Switzerland \& KU Leuven, Belgium \\
              \email{vangool@vision.ee.ethz.ch}
}

\date{Received: date / Accepted: date}

\maketitle

\begin{abstract}
Training deep networks for semantic segmentation requires large amounts of labeled training data, which presents a major challenge in practice, as labeling segmentation masks is a highly labor-intensive process. To address this issue, we present a framework for semi-supervised and domain-adaptive semantic segmentation, which is enhanced by self-supervised monocular depth estimation (SDE) trained only on unlabeled image sequences.

In particular, we utilize SDE as an auxiliary task comprehensively across the entire learning framework: First, we automatically select the most useful samples to be annotated for semantic segmentation based on the correlation of sample diversity and difficulty between SDE and semantic segmentation.
Second, we implement a strong data augmentation by mixing images and labels using the geometry of the scene.
Third, we transfer knowledge from features learned during SDE to semantic segmentation by means of transfer and multi-task learning.
And fourth, we exploit additional labeled synthetic data with Cross-Domain DepthMix and Matching Geometry Sampling to align synthetic and real data.

We validate the proposed model on the Cityscapes dataset, where all four contributions demonstrate significant performance gains, and achieve state-of-the-art results for semi-supervised semantic segmentation as well as for semi\hyp{}supervised domain adaptation. 
In particular, with only 1/30 of the Cityscapes labels, our method achieves 92\% of the fully-supervised baseline performance and even 97\% when exploiting additional data from GTA. The source code is available at 
\url{https://github.com/lhoyer/improving_segmentation_with_selfsupervised_depth}.

\end{abstract}

\section{Introduction}

 Convolutional Neural Networks (CNNs) \citep{lecun1998gradient} have achieved state-of-the-art results for various computer vision tasks including semantic segmentation \citep{long2015fully, chen2017deeplab}. However, training CNNs typically requires large-scale annotated datasets, due to millions of learnable parameters involved. Collecting such training data relies primarily on manual annotation. For semantic segmentation, the process can be particularly costly, due to the required dense annotations. For example, annotating a single image of the Cityscapes dataset took on average 1.5 hours \citep{cordts2016cityscapes}. For the training set, this sums up to 4460 working hours only for the annotation. For more challenging environmental conditions such as fog, snow, or nighttime, the annotation can be even more expensive. For instance, the annotation of one image of the ACDC dataset \citep{sakaridis2021acdc} took 3.3 hours on average.

\begin{figure*}
\centering
\includegraphics[width=0.88\linewidth]{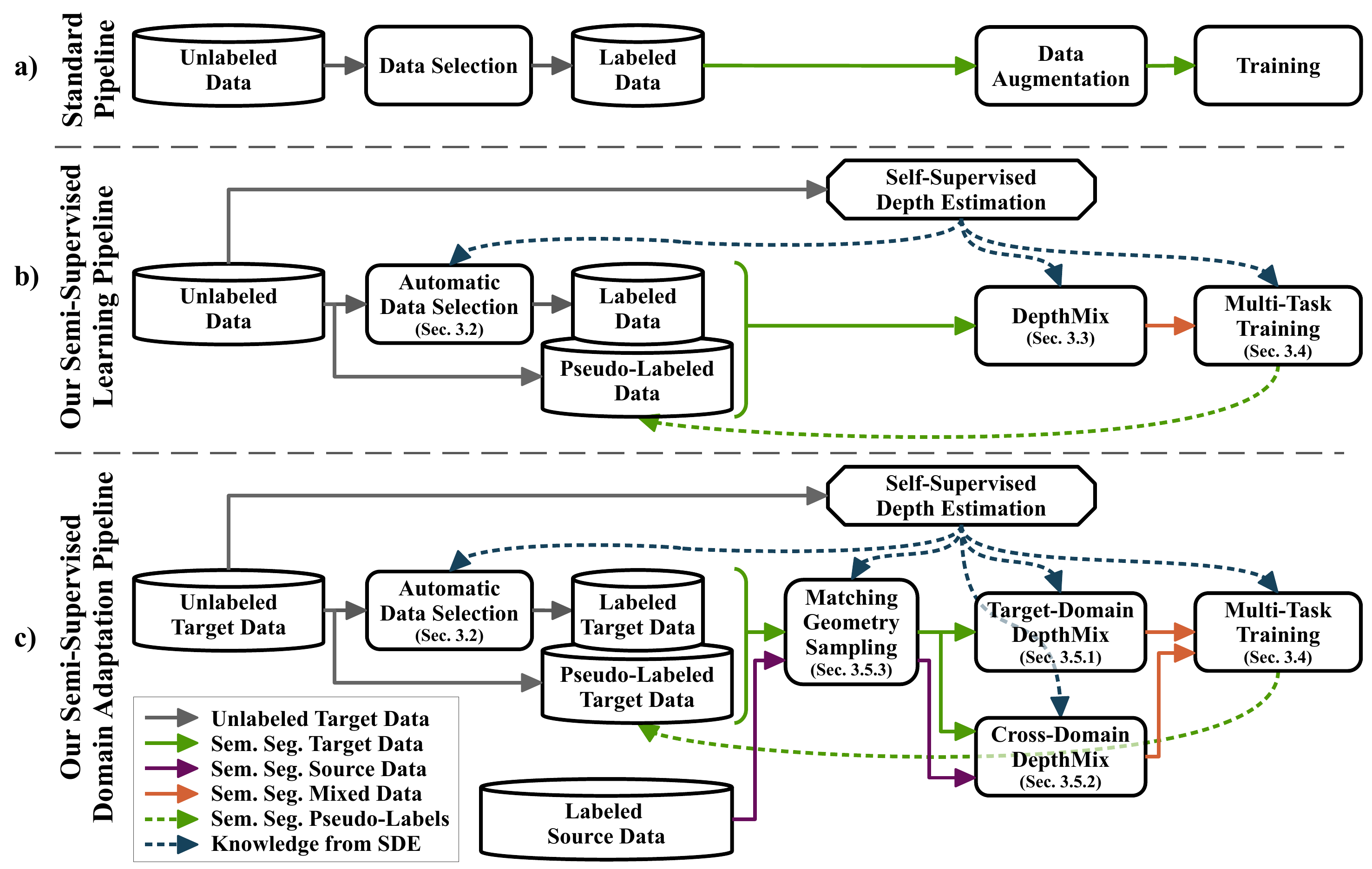}
\caption{Our method utilizes self-supervised depth estimation (SDE) in order to improve the holistic learning process of semantic segmentation. In comparison to the standard learning pipeline (a), we learn SDE from unlabeled image sequences and utilize it to improve the data selection, data augmentation, and training process (b). Further, we extend our framework to semi-supervised domain adaptation (SSDA), where SDE is used to align domains by Matching Geometry Sampling and Cross-Domain DepthMix (c).}
\label{fig:overview}
\end{figure*}

Recently, self-supervised learning \citep{doersch2015unsupervised, gidaris2018unsupervised, he2020momentum} has shown to be a promising replacement for manually labeled data. It aims to learn representations from the structure of unlabeled data, instead of relying on a supervised loss, which requires manual labels. In particular, the principle has successfully been applied in depth estimation for stereo pairs \citep{godard2017unsupervised} or image sequences \citep{zhou2017unsupervised}. 
Additionally, semantic segmentation is known to be tightly coupled with depth. Several works have reported that jointly learning segmentation and \textit{supervised} depth estimation can benefit the performance of both tasks \citep{vandenhende2020revisiting}.
Motivated by these observations, we investigate the question: \textit{How can we leverage self-supervised depth estimation to improve semantic segmentation?}

In this work, we propose to utilize self-supervised monocular depth estimation (SDE) 
\citep{godard2017unsupervised, zhou2017unsupervised, godard2019digging} 
to improve the performance of semantic segmentation and to reduce the number of necessary annotations.
For this purpose, we consider the holistic learning process covering data selection for annotation, data augmentation, domain adaptation, and multi-task learning. For each step, we show how SDE can effectively be utilized to improve the semantic segmentation performance. 
In contrast to most previous works, which only exploit \emph{supervised} depth information during the multi-task learning \citep{vandenhende2020revisiting}, we utilize \emph{self-supervised} depth estimation as an auxiliary task comprehensively across the entire learning pipeline and show that it is critical to effectively improve the segmentation performance.

We apply our framework to the semi-supervised learning (SSL) and the semi-supervised domain adaptation (SSDA) setting. In SSL, only a part of the underlying dataset is labeled for semantic segmentation, while in SSDA additional labeled data from another (often synthetic) domain is provided. 
\autoref{fig:overview} compares the standard learning pipeline (\autoref{fig:overview}a) with our SDE-enhanced framework for SSL (\autoref{fig:overview}b) and our method for SSDA (\autoref{fig:overview}c).

In our SSL framework (see \autoref{fig:overview}b), we utilize SDE learned on unlabeled image sequences, to improve the learning pipeline at three places.

First, we propose an \emph{automatic data selection for annotation}, which selects the most useful samples to be annotated in order to maximize the gain. The selection is iteratively driven by two criteria: \emph{diversity} and \emph{uncertainty}. Both of them are conducted by a novel use of SDE as a proxy task in this context.
While our method follows the active learning cycle (i.e. model training $\rightarrow$ query selection $\rightarrow$ annotation $\rightarrow$ model training) \citep{settles2009active}, it does not require a human in the loop to provide semantic segmentation labels as the human is replaced by a proxy-task SDE oracle. This greatly improves flexibility, scalability, and efficiency, especially considering using crowdsourcing platforms for annotation.

Second, we propose a strong data augmentation strategy, \emph{DepthMix}, which blends images as well as their labels according to the geometry of the scenes obtained from SDE. In comparison to previous methods \citep{yun2019cutmix, olsson2020classmix}, DepthMix explicitly respects the geometric structure of the scenes and generates realistic occlusions as the distance of objects to the camera is considered.

And third, we deploy SDE as an auxiliary task for semantic image segmentation under a transfer learning and multi-task learning framework and show that it noticeably improves the performance of semantic segmentation, especially when semantic supervision is limited. Previous works focus on improving SDE instead of semantic segmentation \citep{chen2019self, guizilini2020semantically} or only consider the special cases of full supervision \citep{klingner2020self} and pretraining \citep{jiang2018self}.

Furthermore, we extend the contributions from SSL to SSDA in order to utilize additional synthetic (source) training data (see \autoref{fig:overview}c). As synthetic data can often be annotated automatically for semantic segmentation, it is a valuable source of supervision and can further reduce the annotation effort for the real (target) data. We demonstrate that the previous contributions are effective for SSDA as well.
In order to better bridge the domain gap between source data and target data, we combine the previous \emph{Target-Domain DepthMix} (i.e. the single-domain DepthMix of our SSL method applied to the target domain) with an additional \emph{Cross-Domain DepthMix}, which mixes samples from the source domain and the target domain. In that way, our framework is able to align the distribution of labeled source data with labeled target data (Cross-Domain DepthMix) and unlabeled target data with labeled target data (Target-Domain DepthMix).
As the geometric distribution of the source domain is not aligned with the target domain and the Cross-Domain DepthMix can suffer from blending samples with different geometric distributions, we further introduce a \emph{Matching Geometry Sampling} based on SDE to better align the camera pose and scene geometry of the source samples with the target domain.

The main advantage of our method is that we can learn from a large base of easily accessible unlabeled image sequences and utilize the learned knowledge from SDE to improve semantic segmentation performance over the entire training process. This largely alleviates the need for expensive semantic segmentation annotations. In our experimental evaluation on Cityscapes \citep{cordts2016cityscapes}, we demonstrate significant performance gains of all four components and improve the previous state of the art for SSL as well as for SSDA by a considerable margin. Importantly, our contributions are complementary and yield even higher improvements when they are combined in a unified framework. Specifically, in an SSL setting, our method achieves 92\% of the fully-supervised model performance with only 1/30 available labels and even slightly outperforms the fully-supervised model with only 1/8 labels. In the SSDA setting with additional supervision from the synthetic GTA5 dataset \citep{richter2016playing}, our method achieves even 97\% of the fully-supervised model performance with only 1/30 of the target labels.

Our contributions summarize as follows:
\begin{itemize}
    \item[(1)] We propose a novel \emph{automatic data selection for annotation} based on SDE to improve the flexibility of active learning for semantic segmentation. It replaces the human annotator with an SDE oracle and lifts the requirement of having a human in the loop of active learning.
    \item[(2)] We propose \emph{DepthMix}, a strong data augmentation strategy based on self-supervised depth estimation, which respects the geometry of the scene. 
    \item[(3)] We utilize SDE as an auxiliary task to exploit depth features learned on unlabeled image sequences to significantly improve the performance of semantic segmentation by transfer and multi-task learning. In combination with (1) and (2), we achieve state-of-the-art results for semi-supervised semantic segmentation on Cityscapes.
    \item[(4)] We propose a novel semi-supervised domain adaptation method, which combines \emph{Target-Domain DepthMix} with \emph{Cross-Domain DepthMix}. Further, \emph{Matching Geometry Sampling} aligns the camera pose and scene geometry during the mixing process towards the target domain. We show that our method achieves state-of-the-art results for SSDA on GTA5$\rightarrow$Cityscapes and Synthia$\rightarrow$Cityscapes.
\end{itemize}

This work is an extension of our IEEE Conference on Computer Vision and Pattern Recognition 2021 paper \citep{hoyer2021three}, which focuses on the contributions (1-3). This article further introduces SSDA utilizing SDE both using the previous contributions for SSL as well as the newly proposed combined Cross-Domain / Target-Domain DepthMix and the Matching Geometry Sampling. Also, we extend the ablation studies, detail the analysis (e.g. by class-wise performance insights and by a class frequency analysis of the data selection), and improve the presentation of the unified SDE-enhanced learning framework.
\section{Related Work}
\label{sec:related_work}

\subsection{Self-Supervised Depth Estimation (SDE)}
\label{sec:sde}

Self-supervised depth estimation (SDE) aims to learn depth estimation from the geometric relations of stereo image pairs \citep{garg2016unsupervised,godard2017unsupervised} or monocular videos \citep{zhou2017unsupervised}. Due to the better availability of videos, we use the latter approach, where a neural network estimates the depth and the camera motion of two subsequent images and a photometric loss is computed after a differentiable warping. If the camera intrinsics are not known, their estimation can be incorporated into the learning process as well \citep{gordon2019depth}. 
Follow-up works propose improvements of the loss function \citep{godard2019digging, gonzalezbello2020forget, shu2020feature}, network architecture \citep{wang2019recurrent,  guizilini20203d}, and training scheme \citep{pilzer2018unsupervised, pilzer2019refine, casser2019depth}. To handle dynamic objects, several works \citep{yin2018geonet, chen2019self, ranjan2019competitive} extend the projection model and combine depth estimation with optical flow estimation.

\subsection{Active Learning}

Active learning methods iteratively select the most informative samples to be annotated.
Two main directions for the selection heuristic can be differentiated. 
On the one side, uncertainty-based approaches select samples with a high uncertainty estimated based on, e.g., entropy \citep{hwa2004sample, settles2008analysis} or ensemble disagreement \citep{seung1992query, mccallumzy1998employing}. 
However, this can be prone to querying outliers. 
On the other side, diversity-based approaches select samples, which most increase the diversity of the labeled set \citep{sener2017active, sinha2019variational}.
For segmentation, active learning is typically based on uncertainty measures such as MC dropout \citep{gal2016dropout, yang2017suggestive, mackowiak2018cereals},
entropy \citep{kasarla2019region, xie2020deal}, or multi-view consistency \citep{siddiqui2020viewal}.
In contrast to these works, we perform automatic data selection for annotation by replacing the human with an SDE model as oracle. Therefore, we do not require human-in-the-loop annotation during the active learning cycle. Previous works performing data selection without a human in the loop are restricted to shallow models \citep{yu2006active, nie2013early, li2018joint}, classification with low-dimensional inputs \citep{li2020deep}, or do not perform an iterative data selection \citep{zheng2019biomedical} to dynamically adapt to the uncertainty of the model trained on the currently labeled set.

\subsection{Semi-Supervised Semantic Segmentation}

Semi-supervised semantic segmentation makes use of additional unlabeled data during training. 
An early line of work \citep{souly2017semi, hung2018adversarial, mittal2019semi} utilizes generative adversarial networks \citep{goodfellow2014generative} in order to include the unlabeled data into the training.

Another increasingly popular direction is self-training with pseudo-labels \citep{lee2013pseudo}, which alternates between prediction of pseudo-labels for unlabeled data and model retraining on the (pseudo-)labeled data. To construct the pseudo-labels, a popular approach is the mean teacher framework \citep{tarvainen2017mean}. It constructs the teacher network for pseudo-label generation from the exponential moving average of the weights of the student network.
In order to avoid lazily mimicking the teacher's predictions and resisting updates, ATSO \citep{huo2021atso} splits the dataset into two parts, trains a model on each, and uses the model trained on one dataset to label the other. Similarly, CPS \citep{chen2021semia} utilizes two networks with different initialization to generate the pseudo-labels for each other.
Further extensions for self-training include training an additional error correction network \citep{mendel2020semi} and dynamically weighing pseudo-labels according to the agreement between two models \citep{fenga2020dmt}.

Self-training is often combined with consistency training, where perturbations are applied to unlabeled images or their intermediate features and a loss term enforces consistency of the predictions. For instance, \citet{ouali2020semi} study perturbation of encoder features, \citet{lai2021semi} enforce consistency of overlapping regions of two crops of the same image with different context, and \citet{sohn2020fixmatch} train the model on strongly augmented images while the pseudo-labels were generated only with weak augmentation. This general framework is extended by several strong augmentation strategies designed for semantic segmentation.
CutMix \citep{yun2019cutmix, french2019consistency} mixes crops from images and their pseudo-labels to generate additional training data, ClassMix \citep{olsson2020classmix} uses class segments of pseudo-labels to build the mix mask, and \citet{dvornik2019importance} paste instance crops into matching context regions of other images.
Our proposed DepthMix module is inspired by these methods but it further respects the geometry of the scene when mixing samples in order to produce realistic occlusions.

\subsection{Multi-Task Learning of Semantic Segmentation and Self-Supervised Depth Estimation}
\label{sec:segmentation_and_sde}

Jointly learning semantic segmentation and SDE was studied in previous works with the goal of improving \textit{depth} estimation. 
Several works \citep{ramirez2018geometry, jiao2018look, yang2018segstereo, chen2019towards, klingner2020self} learn both tasks jointly with a single network. 
Another line of work \citep{casser2019depth, guizilini2020semantically, jiang2019sense} distills knowledge from a teacher semantic segmentation network to guide SDE. 
To further utilize coherence between semantic segmentation and SDE, \citet{ramirez2018geometry} and \citet{chen2019towards} propose a loss term to encourage spatial proximity between depth discontinuities and segmentation contours. 
As moving objects break the static world assumption of the SDE warping process, \citet{casser2019depth} and \citet{klingner2020self} incorporate dynamic object segmentations into the SDE loss calculation.

In contrast to these works, we do not aim to improve SDE but rather semi-supervised semantic segmentation. 
The closest to our approach are \citet{jiang2018self}, \citet{novoselboosting}, and \citet{klingner2020self}. 
\citet{jiang2018self} utilize relative depth computed from optical flow to replace ImageNet pretraining for semantic segmentation.
In contrast, we additionally study multi-task learning of SDE and semantic segmentation and show that combining SDE with ImageNet features can further boost performance.
\citet{novoselboosting} and \citet{klingner2020self} improve the semantic segmentation performance by jointly learning with SDE. However, they focus on the fully-supervised setting, while our work explicitly addresses the challenges of semi-supervised semantic segmentation by using the depth estimates to generate additional training data and an automatic data selection mechanism based on SDE.
Another work \citep{klingner2020improved} supports the usefulness of SDE by improving the robustness of semantic segmentation.

\subsection{Domain Adaptive Semantic Image Segmentation}

A special kind of semi-supervised semantic segmentation is domain adaptation, where the unlabeled and labeled data originate from different domains. Different domains can be, for instance, real and synthetic data \citep{hoffman2016fcns} or data captured under different conditions such as daytime/nighttime \citep{dai2018dark} or weather \citep{sakaridis2018semantic}. Further, it can be distinguished between unsupervised domain adaptation (UDA), if no labeled target data is available, and semi-supervised domain adaptation (SSDA), if a small number of annotations is available for the target domain.

For semantic segmentation, the better-studied scenario is UDA. In order to overcome the domain shift from the source to the target domain, adversarial training can be applied to the input \citep{hoffman2018cycada}, feature \citep{tsai2018learning}, and output space \citep{tsai2018learning, vu2019advent}. Also, non-adversarial input style transfer methods can be utilized \citep{yang2020fda, kim2020learning}.
An increasingly popular approach for UDA is self-training \citep{chapelle2009semi}, where high-confidence predictions of a trained model are used to generate pseudo-labels for unlabeled data to iteratively improve the model \citep{zou2018unsupervised, wei2018revisiting}. DACS \citep{tranheden2021dacs} shows that ClassMix \citep{olsson2020classmix} can also be applied to images from different domains. In contrast to DACS, our method uses the proposed DepthMix strategy, which respects the geometry of the scene during mixing to avoid geometric artifacts, and it combines Cross-Domain DepthMix with Target-Domain DepthMix for effective SSDA. Furthermore, we propose Matching Geometry Sampling to align the scene geometry and camera perspective for Cross-Domain DepthMix. A similar approach has been developed by \citet{li2020content} by sampling images from the source domain with a similar semantic layout as the target domain. However, they do not perform data mixing, do not consider the geometry of the scene, and rely on the generalization from the semantic segmentation network trained on the source domain to the target domain in order to perform the semantic layout matching. As we use SDE, which can be trained on both the source and the target domain, our Matching Geometry Sampling lifts this assumption.
Further self-training extensions include curriculum learning \citep{dai2018dark,zhang2019curriculum, lian2019constructing}, refining pseudo-labels using uncertainties \citep{zheng2021rectifying}, augmentation consistency \citep{araslanov2021self}, and class prototypes \citep{zhang2021prototypical}.

In contrast to UDA, semi-supervised domain adaptation (SSDA), where a few annotations are also available for the target domain, is less studied. \citet{kalluri2019universal} propose a framework with a domain-shared encoder and a domain-specific decoder with additional entropy minimization in a separate embedding space. \citet{wang2020alleviating} extend adversarial domain alignment from UDA \citep{tsai2018learning} and utilizes the additional target labels by applying feature-level adversarial domain alignment between labeled source and labeled target samples. For that, a spatial and a class-wise discriminator are introduced to mitigate inter-class confusions. To produce a better feature representation, \citet{alonso2021semi} extend self-training with a student-teacher framework by contrastive learning \citep{hadsell2006dimensionality}.
Concurrent to our work, \citet{chen2021semi} propose to train one teacher model on domain-mixed batches and one teacher model on CutMix \citep{yun2019cutmix, french2019consistency} batches. A student model is trained on an ensemble of the two teachers and iterative pseudo-labeling is applied to the training of teachers and students. In contrast to these works, our method requires neither sensitive adversarial training nor costly ensemble training. Also, instead of CutMix, we utilize our DepthMix algorithm, which produces geometrically valid synthesized samples. Further, we propose a combined Cross-Domain and Target-Domain DepthMix as well as a Matching Geometry Sampling, which leads to more effective SSDA.

\subsection{Auxiliary Depth Estimation for Domain Adaptation}

For UDA, depth estimates can be another valuable source of supervision to align the domains. For that purpose, SPIGAN \citep{lee2018spigan} and DADA \citep{vu2019dada} extend domain adversarial learning with privileged depth information from the source domain. GIO-Ada \citep{chen2019learning} additionally utilizes the depth information for input style transfer. By providing depth information from the target domain as well, ATDT \citep{ramirez2019learning} learns a bottleneck feature domain transfer network with depth supervision on both domains, which generalizes to semantic segmentation.
In contrast to our work, these approaches require depth ground truth, which can be difficult to acquire.

Concurrently to this work, SDE has been studied as an auxiliary task for \textit{unsupervised} domain adaptation. \citet{guizilini2021geometric} utilize multi-task learning of semantic segmentation and SDE to learn a more domain-invariant representation. Instead of applying the view synthesis loss from SDE directly, \citet{wang2021domain} use depth pseudo-labels from an SDE teacher network to learn depth estimation and semantic segmentation in a multi-tasking framework. To better transfer knowledge between both domains and tasks, the correlation of depth and semantic segmentation features is explicitly transferred from the source to the target domain and the depth adaptation difficulty is transferred to semantic segmentation to weigh the trust in the semantic segmentation pseudo-labels.  
Using (self-supervised) depth estimation for \emph{semi-supervised} domain adaptation, however, has not been studied so far.

\section{Methods}
\label{sec:methods}

In this chapter, we present our four approaches to improve the performance of semantic segmentation with self-supervised depth estimation (SDE). They focus on four different aspects of the training process, covering data selection for annotation, data augmentation, multi-task learning, and domain adaptation. Given $N$ images and $M$ image sequences from the same domain, our first method, \emph{automatic data selection for annotation}, uses SDE learned on the $M$ (unlabeled) sequences to select $N_A$ images out of the $N$ images for human annotation (see \autoref{sec:data_selection}). Our second approach, termed \emph{DepthMix}, leverages the learned SDE to create geometrically-sound `virtual' training samples from pairs of labeled images and their annotations (see \autoref{sec:methods_depthmix}). Our third method learns semantic segmentation with SDE as an auxiliary task under a multi-tasking framework (see \autoref{sec:methods_mtl}). The learning is reinforced by a multi-task pretraining process combining SDE with image classification. 
And fourth, we extend our method to semi-supervised domain adaptation (SSDA) in order to utilize additional synthetic data, which has a low labeling effort (see \autoref{sec:methods_ssda}). To address the domain gap, we propose a combined \emph{Cross-Domain} and \emph{Target-Domain DepthMix} strategy, which is enhanced by \emph{Matching Geometry Sampling}.

\subsection{Self-Supervised Depth Estimation (SDE)}

For self-supervised depth estimation (SDE), we follow the method of \citet{godard2019digging}, which we briefly introduce in the following. We first train a depth estimation network to predict the depth of a target image and a pose estimation network to estimate the camera motion from the target image and the source image. Depth and pose are used to produce a differentiable warping to transform the source image into the target image. The photometric error between the target image and multiple warped source frames is combined by a pixel-wise minimum. Besides, stationary pixels are masked out and an edge-aware depth smoothness term is applied resulting in the final SDE loss $L_D$. We refer the reader to the original paper \citep{godard2019digging} for more details.

\subsection{Automatic Data Selection for Annotation}
\label{sec:data_selection}

\begin{figure}
\centering
\vspace{0.1cm}
\includegraphics[width=1\linewidth]{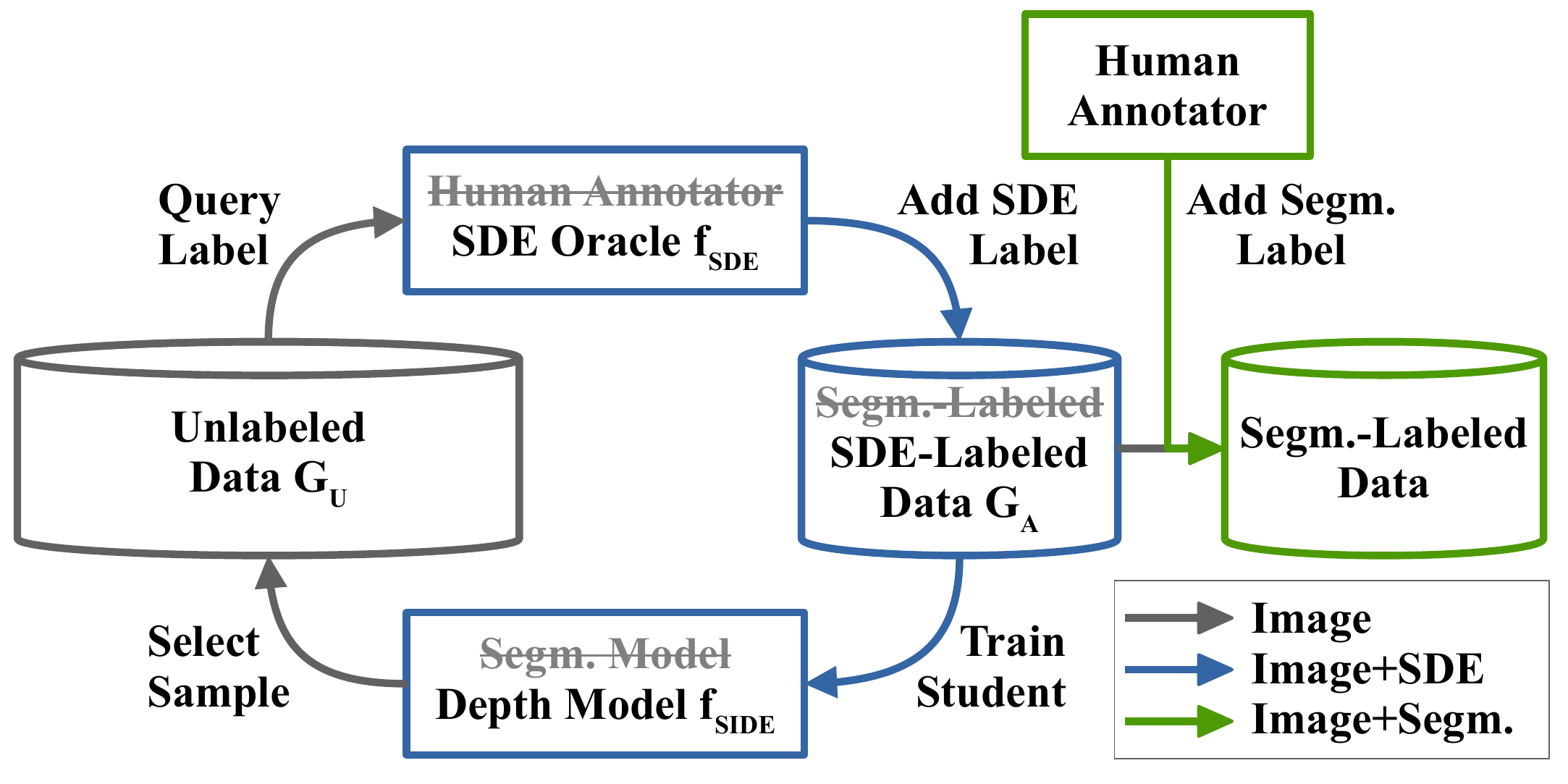}
\caption{The automatic data selection for annotation process selects the most useful samples from the set of unlabeled data $\mathcal{G}_U$ to be annotated. In contrast to active learning, the human annotator is replaced by an SDE oracle, and the samples are selected according to depth estimation as proxy-task. This lifts the requirement of a human in the loop. Samples are selected according to SDE feature diversity (\autoref{sec:methods_diversity_sampling}) and depth student uncertainty (\autoref{sec:methods_uncertainty_sampling}).}
\label{fig:dataselection}
\end{figure}

We use SDE as a proxy task for selecting $N_A$ samples out of a set of $N$ unlabeled samples for a human to create semantic segmentation labels.
The selection is conducted progressively in multiple steps, similar to the standard active learning cycle (model training $\rightarrow$ query selection $\rightarrow$ annotation $\rightarrow$ model training). However, our data selection is fully automatic and does not require a human in the loop as the annotation is done by a proxy-task SDE oracle as visualized in \autoref{fig:dataselection}.

Let's denote by $\mathcal{G}$, $\mathcal{G}_A$, and $\mathcal{G}_U$, the whole image set, the selected subset for annotation, and the unselected subset.   
Initially, we have $\mathcal{G}_A=\emptyset$ and $\mathcal{G}_U=\mathcal{G}$. The selection is driven by two criteria: \emph{diversity} and \emph{uncertainty}. Diversity sampling encourages the selected images to be diverse and cover different scenes. Uncertainty sampling favors adding unlabeled images that are near a decision boundary (with high uncertainties) of the model trained on the current $\mathcal{G}_A$. 
For uncertainty sampling, we need to train and update the model with $\mathcal{G}_A$. It is inefficient to repeat this every time a new image is added. For the sake of efficiency, we divide the selection into $T$ steps and only train the model $T$ times. In each step $t$, $n_t$ images are selected and moved from $\mathcal{G}_U$ to $\mathcal{G}_A$, so we have $\sum_{t=1}^T n_t = N_A$. After each step $t$, a model is trained on $\mathcal{G}_A$ and evaluated on $\mathcal{G}_U$ to get updated uncertainties for step $t+1$. 

\begin{algorithm}[tb]
\caption{Automatic Data Selection}
\label{alg:label_selection}
\begin{algorithmic}[1]
\State $t=1$
\State $i \gets \text{uniform}(1,N)$ 
\State $\mathcal{G}_A = \{I_i\} \text{ and } \mathcal{G}_U = \mathcal{G}_U \setminus \{I_i\}$ 
\For{$k=2$ {\bfseries to} $N_A$}
\If{ $k==\sum_{t'=1}^tn_{t'}$}
\State Train depth student $\Phi_\text{SIDE}$ on $\mathcal{G}_A$
\State Calculate $E(i) \text{ } \forall I_i \in \mathcal{G}_U$
\State $t = t + 1$
 \EndIf
 \If{$t==1$} 
     \State Obtain index $i$ according to \autoref{eq:diversity} 
 \Else 
      \State Obtain index $i$ according to \autoref{eq:uncertainty:diversity} 
 \EndIf
    \State $\mathcal{G}_A = \mathcal{G}_A \cup \{I_i\} \text{ and } \mathcal{G}_U = \mathcal{G}_U \setminus \{I_i\}$
\EndFor
\end{algorithmic}
\end{algorithm}

\subsubsection{Diversity Sampling}
\label{sec:methods_diversity_sampling}

To ensure that the chosen annotated samples are diverse enough to represent the entire dataset well, we use an iterative farthest point sampling based on the L2 distance over features $\Phi^{\text{SDE}}$ computed by an intermediate layer of the SDE network.
At step $t$, for each of the $n_t$ samples, we choose the one in $\mathcal{G}_U$ with the largest distance to the current annotation set $\mathcal{G}_A$.
The set of selected samples $\mathcal{G}_A$ is iteratively extended by moving one image at a time from $\mathcal{G}_U$ to $\mathcal{G}_A$ until the $n_t$ images are collected:
\begin{equation}
\label{eq:set:enlarge}
    \mathcal{G}_U  = \mathcal{G}_U  \setminus \{I_i\} \text{ and } \mathcal{G}_A =  \mathcal{G}_A \cup \{I_i\}\,,
\end{equation}
\begin{equation}
\label{eq:diversity}
     i = \argmax_{I_i\in \mathcal{G}_U} \min_{I_j\in \mathcal{G}_A} ||\Phi^{\text{SDE}}_i - \Phi^\text{SDE}_j||_2\,.
\end{equation}

\subsubsection{Uncertainty Sampling}
\label{sec:methods_uncertainty_sampling}

While diversity sampling is able to select diverse new samples, it is unaware of the uncertainties of a semantic segmentation model over these samples. Our uncertainty sampling aims to select difficult samples, i.e., samples in $\mathcal{G}_U$ that the model trained on the current $\mathcal{G}_A$ cannot handle well.
In order to train this model, active learning typically uses a human-in-the-loop strategy to add annotations for selected samples. In this work, we use a proxy task based on self-supervised annotations, which can run automatically, to make the method more flexible and efficient.
Since our target task is single-image semantic segmentation, we choose to use single-image depth estimation (SIDE) as the proxy task. Importantly, due to our SDE framework, depth pseudo-labels are available for $\mathcal{G}$. Using these pseudo-labels, we train a SIDE method on $\mathcal{G}_A$ and measure the uncertainty of its depth predictions on $\mathcal{G}_U$. Due to the high correlation of single-image semantic segmentation and SIDE, the generated uncertainties are informative and can be used to guide our sampling procedure. As the depth student model is trained only on $\mathcal{G}_A$, it can specifically approximate the difficulty of candidate samples with respect to the already selected samples in $\mathcal{G}_A$. The student is trained from scratch in each step $t$, instead of being fine-tuned from $t-1$, to avoid getting stuck in the previous local minimum. Note that the SDE method is trained on a much larger unlabeled dataset, i.e., the $M$ image sequences, and can provide good guidance for the SIDE method.

In particular, the uncertainty is signaled by the disparity error between the student network $f_{\text{SIDE}}$ and the teacher network $f_{\text{SDE}}$ in the log-scale space under L1 distance: 
\begin{equation}
\label{eq:uncertainty}
    E(i) = || \log(1 + f_{\text{SDE}}(I_i)) - \log(1 + f_{\text{SIDE}}(I_i))||_1\,.
\end{equation}
As the disparity difference of far-away objects is small, the log-scale is used to avoid the loss being dominated by close-range objects. This criterion can be added into \autoref{eq:diversity} to also select samples with higher uncertainties for the dataset update in \autoref{eq:set:enlarge}: 
\begin{equation}
    \label{eq:uncertainty:diversity}
     i = \argmax_{I_i\in \mathcal{G}_U} \min_{I_j\in \mathcal{G}_A} ||\Phi^{\text{SDE}}_i - \Phi^\text{SDE}_j||_2 + \lambda_{\text{E}}E(i)\,,
\end{equation}
where $\lambda_E$ is a parameter to balance the contribution of the two terms. For diversity sampling, we still use SDE features instead of SIDE student features as SDE is trained on the entire dataset, which provides better features for diversity estimation.
When $n_t$ images have been selected according to \autoref{eq:set:enlarge} and \autoref{eq:uncertainty:diversity} at step $t$, a new SIDE model will be trained on the current $\mathcal{G}_A$ in order to continue further.
As presented previously, our selection proceeds progressively in $T$ steps until we collect all $N_A$ images. The algorithm of this selection is summarized in \autoref{alg:label_selection}, where $\sum_{t'=1}^tn_{t'}$ describes the desired size of $\mathcal{G}_A$ at the end of step $t$.


\subsection{DepthMix Data Augmentation}
\label{sec:methods_depthmix}

Inspired by the recent success of data augmentation approaches that mixup pairs of images and their (pseudo) labels to generate more training samples for semi-supervised semantic segmentation \citep{yun2019cutmix, french2019consistency, olsson2020classmix}, we propose an algorithm, termed DepthMix, to utilize self-supervised depth estimates to maintain the integrity of the scene structure during mixing.

\begin{figure}
\centering
\vspace{0.1cm}
\includegraphics[width=\linewidth]{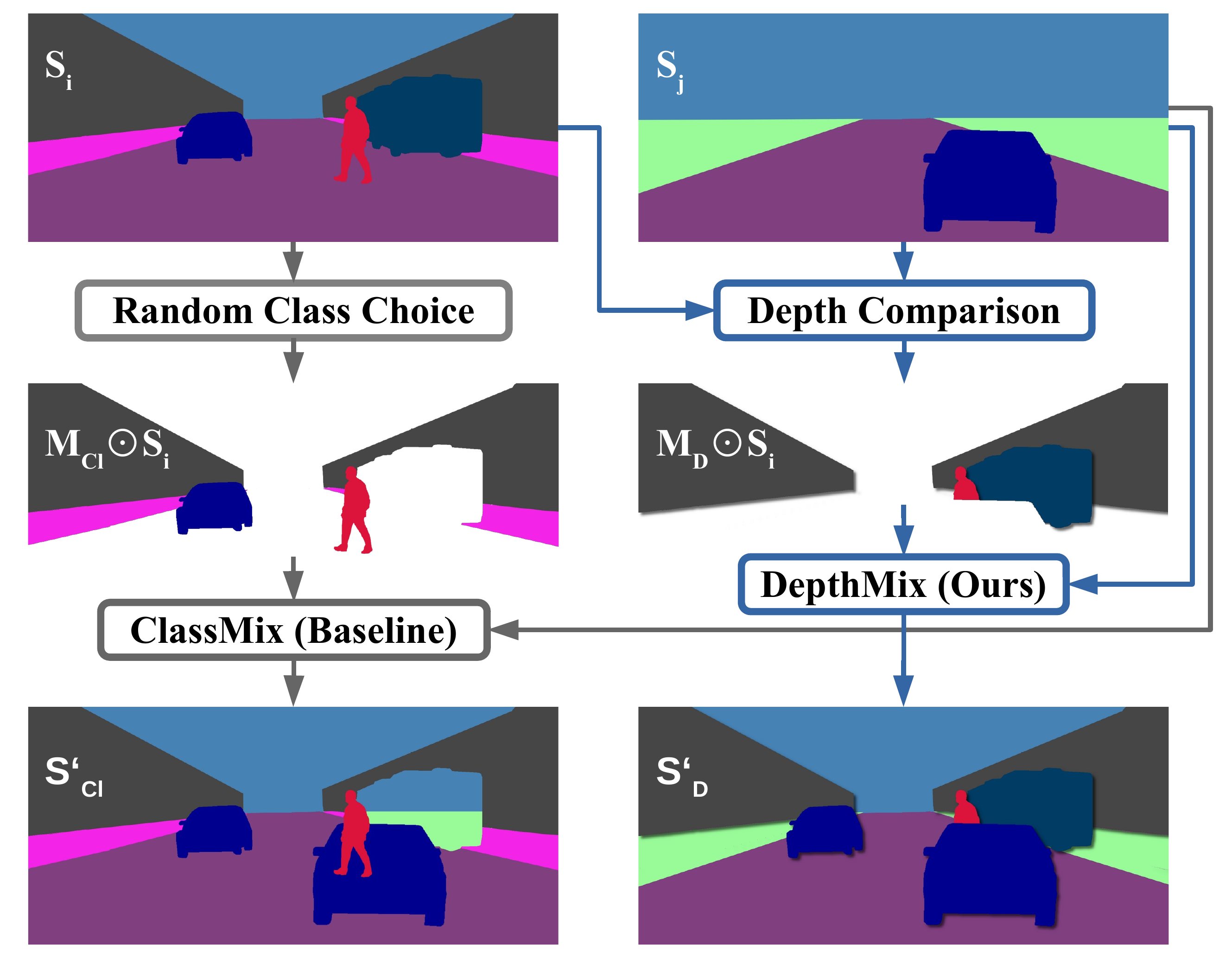}
\caption{Concept of the proposed DepthMix data augmentation (refer to \autoref{sec:methods_depthmix}) and its baseline ClassMix \citep{olsson2020classmix} shown for the mixing of the semantic segmentation labels. By utilizing SDE, DepthMix mitigates geometric artifacts such as missing occluders (bus-shaped hole in the building) or missing occlusion (legs of the person). The corresponding images are mixed in the same way.}
\label{fig:depthmix}
\end{figure}

Given two images $I_i$ and $I_j$ of the same size, we would like to copy some regions from $I_i$ and paste them directly into $I_j$ to get a virtual sample $I^\prime$. The copied regions are indicated by a binary mask $M$, which has the same size as the two images. The image creation is done as
\begin{equation}
    I^\prime = M \odot I_i + (1 - M) \odot I_j\,,
    \label{eq:mix}
\end{equation}
where $\odot$ denotes the element-wise product. 
The semantic segmentation labels of the two images $S_i$ and $S_j$ are mixed up with the same mask $M$ to generate the corresponding mixed semantic segmentation
\begin{equation}
    S^\prime = M \odot S_i + (1 - M) \odot S_j\,.
    \label{eq:mix_s}
\end{equation}
The mixing can be applied to labeled data and unlabeled data using human ground truths or pseudo-labels, respectively.
Existing methods generate this mask $M$ in different ways, e.g., randomly sampled rectangular regions \citep{yun2019cutmix, french2019consistency} or randomly selected class segments \citep{olsson2020classmix}.
In those methods, the structure of the scene is not considered and foreground and background are not distinguished.
We find images synthesized by these methods often violate the geometric relationships between objects. For instance, a distant object can be copied onto a close-range object or only unoccluded parts of mid-range objects are copied onto the other image. Imagine how strange it is to see a pedestrian standing on top of a car or to see the sky through a hole in a building (just as shown in \autoref{fig:depthmix} left).

Our DepthMix is designed to mitigate this issue. It uses the self-supervised depth estimates $\hat{D}i$ and $\hat{D}j$ of the two images to generate the mask $M$, which respects the notion of geometry. It is implemented by selecting only pixels from $I_i$ whose depth values are smaller than the depth values of the pixels at the same locations in $I_j$: 
\begin{equation}
    M(a,b) = \left\{ 
    \begin{array}{rl} 
    1 & \text{if } \hat{D}_i(a,b) < \hat{D}_j(a,b) + \epsilon \\
    0 & \text{otherwise } 
     \end{array} \right .
\label{eq:depth_mix_mask}
\end{equation}
where $a$ and $b$ are pixel indices, and $\epsilon$ is a small value to avoid conflicts of objects that are naturally at the same depth plane such as road or sky.
By using this $M$, DepthMix respects the depth of objects in both images, such that only closer objects can occlude further-away objects. We illustrate this advantage of DepthMix with an example in \autoref{fig:depthmix}.

In order to further utilize the unlabeled dataset $\mathcal{G}_U$ for DepthMix, we generate pseudo-labels using the mean teacher algorithm \citep{tarvainen2017mean}, which is commonly deployed in SSL \citep{berthelot2019mixmatch, verma2019interpolation, french2019consistency, olsson2020classmix}. For that purpose, an exponential moving average is applied to the weights of the semantic segmentation model $g^S_\theta$ to obtain the weights of the mean teacher $\theta_T$:
\begin{equation}
    \theta'_T = \alpha \theta_T + (1 - \alpha) \theta\,.
\end{equation}
To generate the pseudo-labels, an argmax over the classes $C$ is applied to the prediction of the mean teacher:
\begin{equation}
    S_U = \argmax_{c \in C}(g^S_{\theta_T}(I_U))\,.
\label{eq:S_U}
\end{equation}
The mean teacher can be considered as a temporal ensemble, resulting in stable predictions for the pseudo-labels, while the argmax promotes confident predictions \citep{olsson2020classmix}. 

In order to utilize the pseudo-labels, we apply DepthMix to two samples $(I_i, S_i), (I_j, S_j)$ from the combined labeled and pseudo-labeled data pool $\mathcal{G}_A \cup \mathcal{G}_U$ to produce a mixed training pair $(I', S')$ according to \autoref{eq:mix}. The semantic segmentation network is trained using the cross-entropy of labeled samples $(I_A, S_A)$ and the quality-weighted cross-entropy of mixed samples $(I', S')$:
\begin{equation}
\begin{split}
    L_\mathit{DX} = L_{ce}(g^S_\theta(I_A), S_A) +
    q' L_{ce}(g^S_\theta(I'), S')\,, 
\end{split}
\label{eq:L_SSL_2}
\end{equation}
where $q'$ denotes the estimated quality of the mixed pseudo-label. It is the fraction of pixels exceeding a threshold $\tau$ for the predicted probability of the most confident class $P'$:
\begin{equation}
    q' = \frac{\sum_{a,b} [P'(a, b) > \tau]}{W\cdot H}\,.
\end{equation}
As the DepthMix segmentation $S'$ consists of labels from two images, we calculate $P'$ as the mix of its sources:
\begin{equation}
    P' = M \odot P_i + (1-M) \odot P_j\,,
\end{equation}
where $P$ is the predicted probability of the most confident class for unlabeled images and 1 for labeled images:
\begin{equation}
    P(a,b) = 
    \begin{cases}
        \max_{c \in C}(g^S_{\theta_T}(I)(a,b)),& \text{if } I \in \mathcal{G}_U\\
        1,              & \text{otherwise}
    \end{cases}
\end{equation}

By applying DepthMix to labeled and pseudo-labeled samples, the network is exposed to image regions from both distributions in a single image. This can improve its generalization to the unlabeled data as the context for labeled regions can originate from unlabeled data and vice versa. The improved generalization can lead to better pseudo-labels, which in turn improve the quality of the DepthMix labels.

\subsection{Learning with Auxiliary Self-Supervised Depth Estimation}
\label{sec:methods_mtl}

\begin{figure}
\centering
\vspace{0.1cm}
\includegraphics[width=\linewidth]{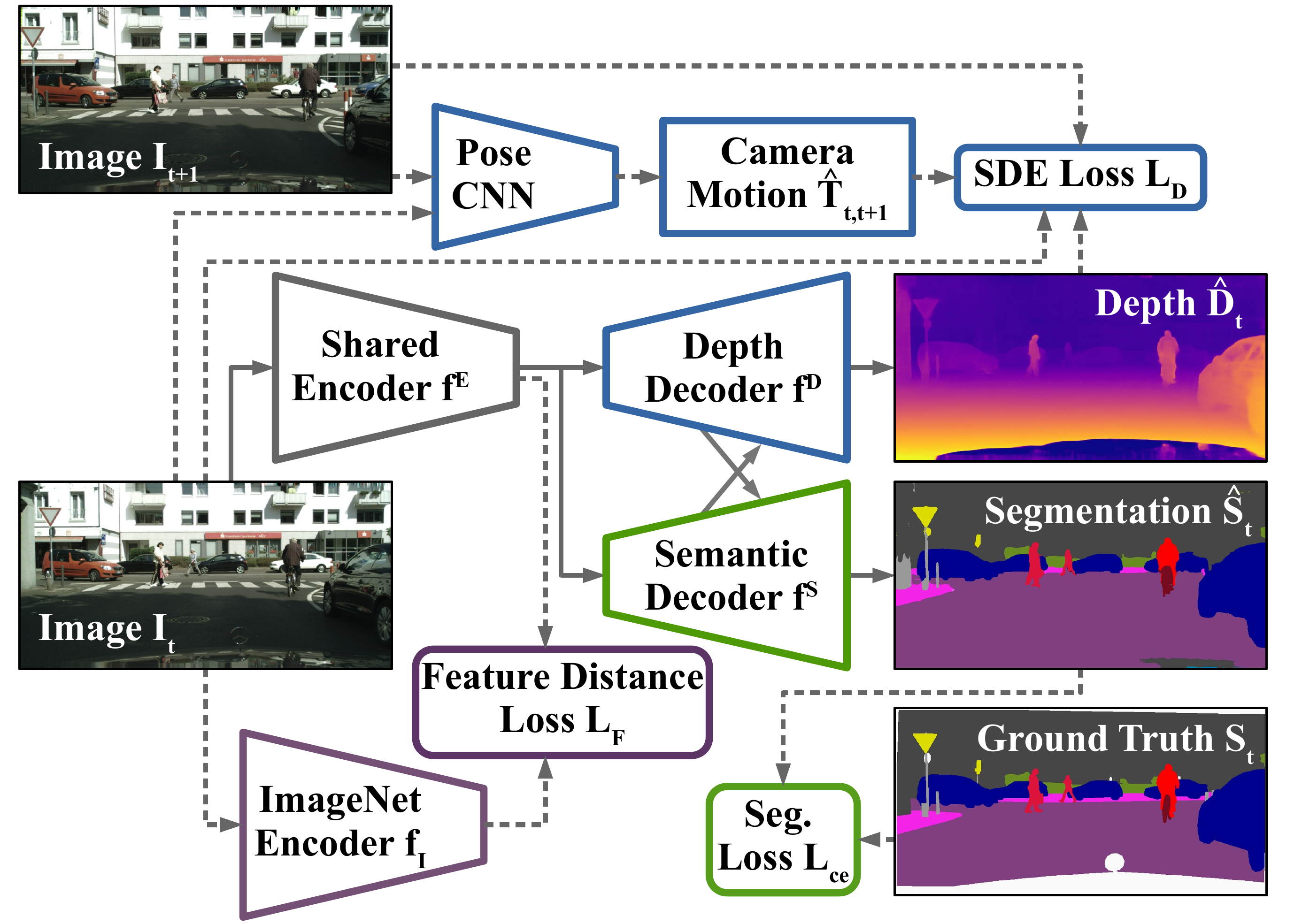}
\vspace{0.2cm}
\caption{Architecture for learning semantic segmentation with SDE as auxiliary task according to \autoref{sec:methods_mtl}. The dashed paths are only used during training and only if image sequences and/or segmentation ground truth are available for a training sample.}
\label{fig:architecture}
\end{figure}

In this section, we utilize features learned by SDE from unlabeled image sequences to improve the performance of semantic segmentation through transfer and multi-task learning. For that purpose, we use a network with a shared encoder $f^E_\theta$, a separate depth decoder $f^D_\theta$, and a separate segmentation decoder $f^S_\theta$ (see \autoref{fig:architecture}). For effective multi-task learning, we use an attention-guided distillation module \citep{xu2018pad} to exchange useful intermediate features between both decoders. The depth branch $g^D_\theta = f^D_\theta \circ f^E_\theta$ is trained using the SDE loss $L_D$ and the segmentation branch $g^S_\theta = f^S_\theta \circ f^E_\theta$ is trained using $L_\mathit{DX}$ (see \autoref{eq:L_SSL_2})
\begin{equation}
    L_\mathit{MTL} = L_D + L_\mathit{DX}\,.
\end{equation}

In order to initialize the pose estimation network and the depth branch $g^D_\theta = f^D_\theta \circ f^E_\theta$ properly, the architecture is first only trained on $M$ unlabeled image sequences for SDE. As a common practice, we initialize the encoder with ImageNet weights as they provide useful semantic features learned during image classification. To avoid forgetting these semantic features during the SDE pretraining, we utilize a feature distance loss between the current bottleneck features $f^{E}_\theta$ and the bottleneck features generated by the encoder with ImageNet weights $f^{E}_{I}$
\begin{equation}
    L_{F} = ||f^{E}_\theta - f^{E}_{I}||_2\,.
\end{equation}
The loss for the depth pretraining is the weighted sum of the SDE loss and the ImageNet feature distance loss
\begin{equation}
    L_{D,\mathit{pretrain}} = L_D + \lambda_{F} L_{F}\,.
\end{equation}

To exploit the features from SDE for semantic segmentation by transfer learning, the weights from SDE $g^{D}_\theta$ are used to initialize the semantic segmentation branch $g^{S}_\theta$.

\subsection{Semi-Supervised Domain Adaptation (SSDA)}
\label{sec:methods_ssda}

Synthetic data can be another valuable source for low-effort semantic segmentation annotations to reduce the number of expensive target labels. In semi-supervised domain adaptation (SSDA), a neural network is trained to solve a task on the real (target) domain while being trained using a limited number of annotated target samples $(I^{trg}_A, S^{trg}_A)$, further unlabeled target images $I^{trg}_U$, and additional annotated data from the synthetic (source) domain $(I^{src}_A, S^{src}_A)$.

Naively, the semantic segmentation network branch $g_\theta^S$ can be trained on the labeled samples from both source and target domain using a pixel-wise cross-entropy loss
\begin{align}
    L_{ce}^{trg} &= L_{ce}(g_\theta^S(I^{trg}_A), S^{trg}_A) \,,
    \label{eq:clean_target_loss}\\
    L_{ce}^{src} &= L_{ce}(g_\theta^S(I^{src}_A), S^{src}_A) \,.
    \label{eq:clean_source_loss}
\end{align}
However, as the labeled data from the target dataset is limited, the vanilla training strategy suffers from the gap between both domains.

In this work, we propose to use SDE to overcome the domain gap of SSDA. Extending the default setup, we augment both the target and the source dataset with self-supervised depth estimates. For that purpose, an SDE network $f_D^{trg}$ is trained on image sequences from the target domain and another SDE network $f_D^{src}$ is trained on image sequences from the source domain. Note that the image sequences can be different from the images labeled for semantic segmentation. After the SDE training, depth pseudo-labels are inferred for the images of the semantic segmentation datasets: $D^{src}_A = f_D^{src}(I^{src}_A)$; $D^{trg}_U = f_D^{trg}(I^{trg}_U)$; $D^{trg}_A = f_D^{trg}(I^{trg}_A)$.
Further, pseudo-labels $S^{trg}_U$ are obtained online according to \autoref{eq:S_U} for the unlabeled target data. The additional depth and semantic segmentation pseudo-labels are added to the SSDA training data. 

\begin{figure}
\centering
\vspace{0.1cm}
\includegraphics[width=1.0\linewidth]{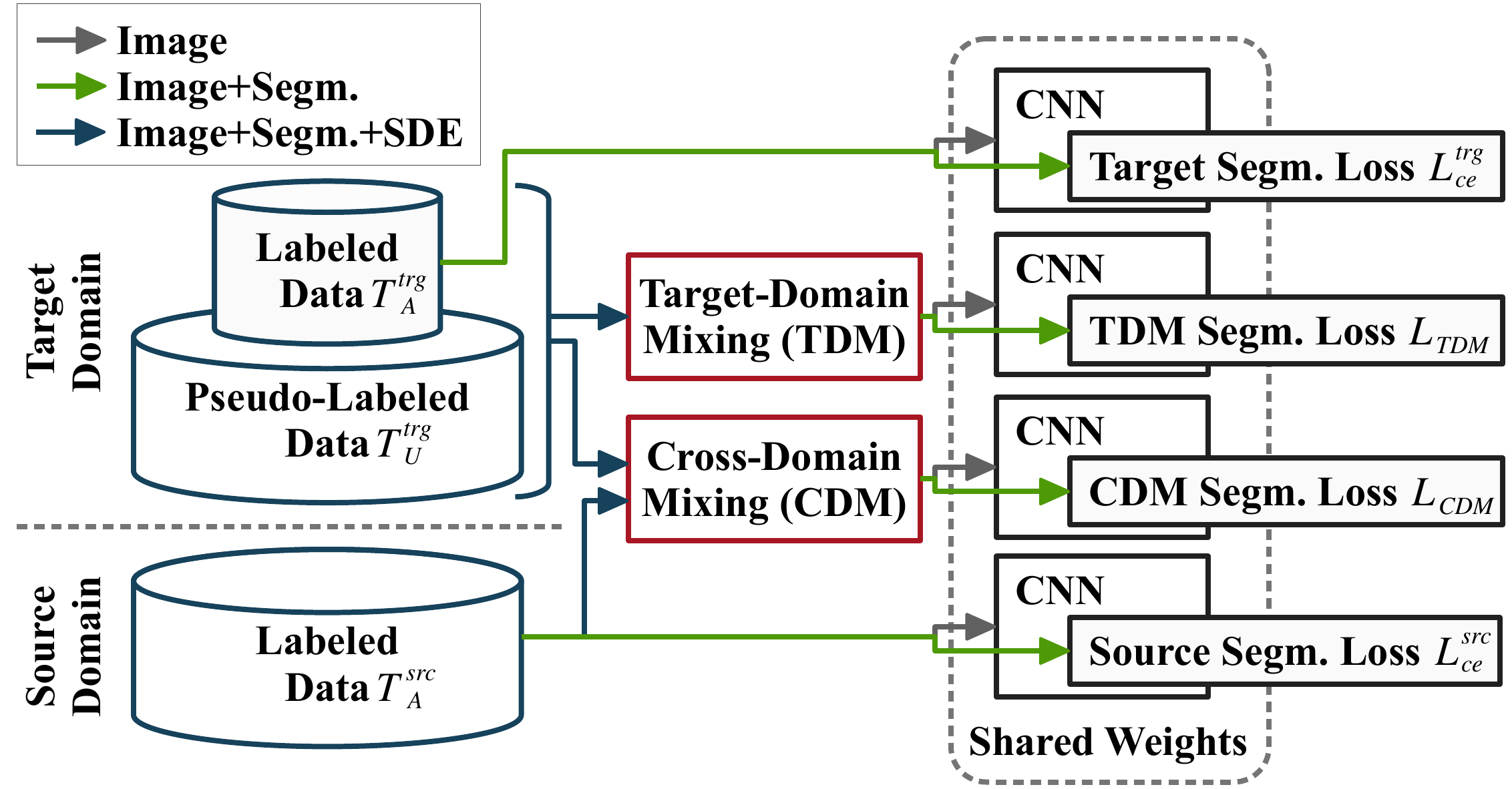}
\caption{Semi-supervised domain adaptation (SSDA) pipeline with Cross-Domain DepthMix (CDM) and Target-Domain DepthMix (TDM). While CDM applies DepthMix to samples from source and target domain to align both domains, TDM mixes labeled and pseudo-labeled samples from the target domain to align labeled and unlabeled target data. The network is trained on clean labeled source data, CDM/TDM data, and clean labeled target data for semantic segmentation. The target semantic segmentation pseudo-labels are obtained online using a mean teacher network.}
\label{fig:ssda}
\end{figure}

Based on this data, we propose a combined Cross\hyp{}Domain and Target-Domain DepthMix in order to facilitate effective self-training across domains as well as across labeled and unlabeled samples, respectively. Further, we enhance the mixing by Matching Geometry Sampling. The training process is visualized in \autoref{fig:ssda} and described in the following.

\subsubsection{Target-Domain DepthMix (TDM)}
\label{sec:methods_TDM}

Target-Domain DepthMix (TDM) applies the DepthMix algorithm to the target dataset. It mixes labeled and unlabeled target samples to improve the generalization from the labeled target to the unlabeled target samples due to the increased variety of objects in different contexts. Therefore, it can favorably affect the quality of the pseudo-labels.
Target-Domain DepthMix uses the same procedure as the single-domain SSL DepthMix described in \autoref{sec:methods_depthmix}. It produces a mixed sample $(I'_\mathit{TDM},\allowbreak S'_\mathit{TDM})$ based on two target samples according to \autoref{eq:mix} -- \ref{eq:depth_mix_mask}.
The segmentation branch of the network is trained using the pixel-wise cross-entropy on the mixed samples
\begin{equation}
    L_\mathit{TDM} = q'_\mathit{TDM} L_{ce}(g_\theta^S(I'_\mathit{TDM}), S'_\mathit{TDM})\,,
\end{equation}
where $q'_\mathit{TDM}$ weighs the loss according to the certainty of the pseudo-label as described in \autoref{sec:methods_depthmix}.

Mixing within a domain is only applied to the target domain and not to the source domain because the mixing serves the purpose of better generalization from labeled to unlabeled samples during the self-training. The source domain already contains many labeled samples. Therefore, self-training augmented by mixing is not necessary.

\subsubsection{Cross-Domain DepthMix (CDM)}
\label{sec:methods_CDM}

As there is only a small number of labeled samples available for the target domain, the trained network will still suffer from the gap between the source and target domain. To further align the domains during training, we propose Cross-Domain DepthMix, which mixes samples from both domains. This allows the network to better generalize across domains as both domains are present within each image.

Cross-Domain DepthMix utilizes one target sample and one source sample. If the target image is unlabeled, a pseudo-label is generated according to \autoref{eq:S_U}. The samples are mixed according to \autoref{eq:mix} -- \ref{eq:depth_mix_mask} to generate the cross-domain mixed sample $(I'_\mathit{CDM}, S'_\mathit{CDM})$. The segmentation branch of the network is trained using the pixel-wise cross-entropy on the mixed samples
\begin{equation}
    L_\mathit{CDM} = q'_\mathit{CDM} L_{ce}(g_\theta^S(I'_\mathit{CDM}), S'_\mathit{CDM})\,,
    \label{eq:CDM}
\end{equation}
where $q'_{CDM}$ weighs the loss according to the certainty of the pseudo-label as described in \autoref{sec:methods_depthmix}.

The final SSDA loss combines all four segmentation losses as well as the SDE loss on the target domain
\begin{equation}
    L_{SSDA} = L_{ce}^{trg} + L_{ce}^{src} + L_\mathit{CDM} + L_\mathit{TDM} + L_D^{trg}\,,
    \label{eq:L_SSDA}
\end{equation}
where the loss components are weighted equally.

\subsubsection{Matching Geometry Data Sampling}
\label{sec:methods_mg}

\begin{figure}
\centering
\vspace{0.1cm}
\includegraphics[width=\linewidth]{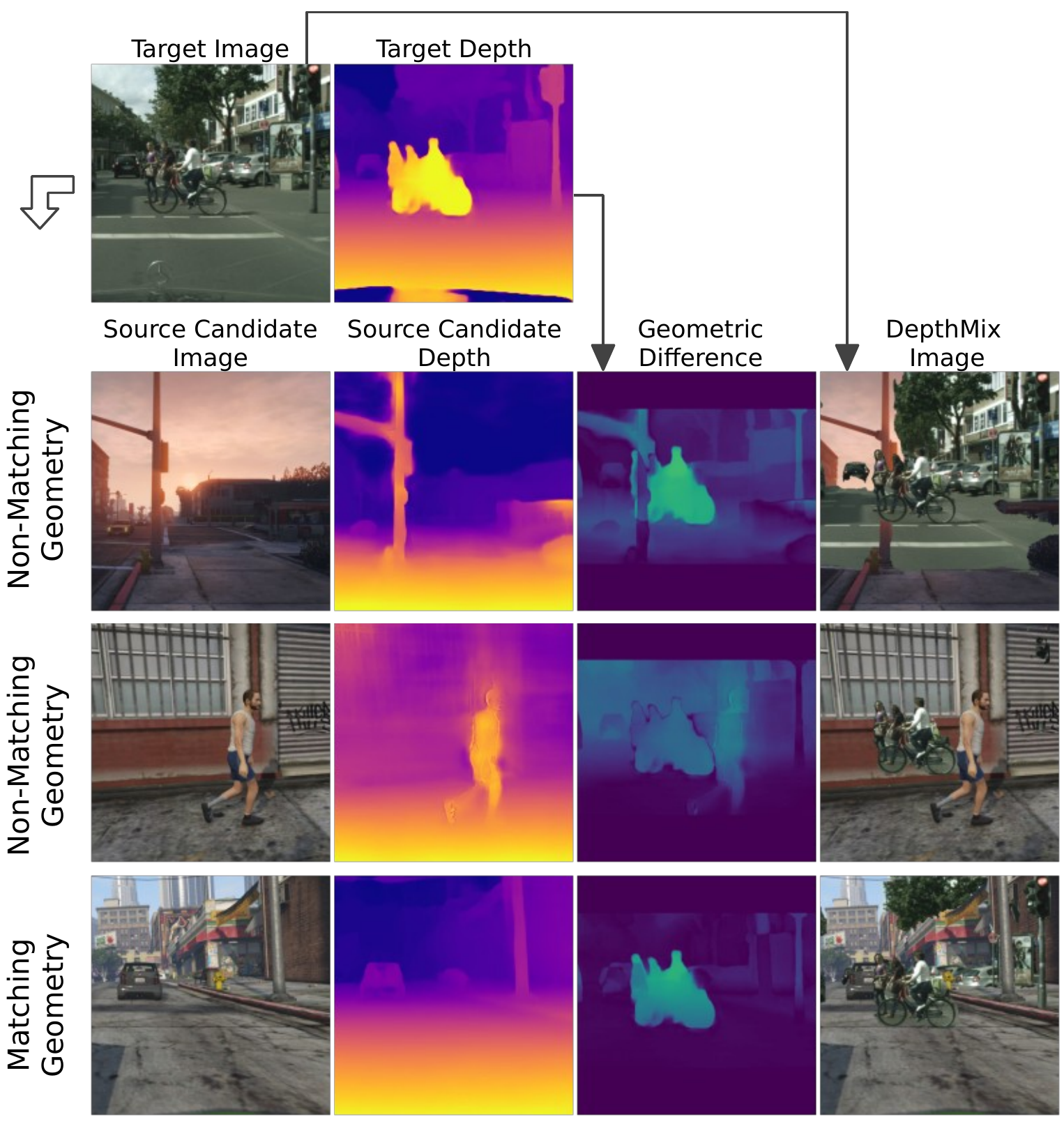}
\caption{Examples of the geometric domain gap and the Matching Geometry Sampling. Images and their SDE are shown for the target (first row) and the source domain (remaining rows). Some samples from the source domain (second and third row) have a different depth distribution compared to the target domain, which results in unrealistic DepthMix images (last column). Matching geometry sampling avoids sampling those domain pairs by selecting pairs with a small geometric difference (fourth row).}
\label{fig:mg_showcase}
\end{figure}

For samples from two different domains, the camera pose can differ between the domains as can be seen in the first three rows of \autoref{fig:mg_showcase}. The geometric distribution difference between domains can impede the transfer of knowledge from the source to the target domain. For example, GTA contains samples from the view of a pedestrian while all Cityscapes samples are recorded from a front-facing camera of a car. This leads to different camera perspectives, which can result in unrealistic mixed samples such as a car ``flying" in the sky (second row), or samples out of the target distribution such as images captured right in front of a building (third row).

We address this problem by sampling image pairs from the source and the target domain with a similar geometry with respect to the camera. The sampling is guided by the target geometry, which allows us to better match the geometric target distribution with mixed images.
We define the geometric difference $G(i,j)$ of two samples $i$ and $j$ as the L1 distance of the log-scale disparity (inverse depth) estimates in camera space
\begin{equation}
    G(i,j) = ||\log(1+\frac{1}{D_i}) - \log(1+\frac{1}{D}_j)||_1\,,
    \label{eq:geometric_difference}
\end{equation}
which corresponds to the metric used for the uncertainty sampling of our automatic data selection in \autoref{eq:uncertainty}. When calculating the geometric difference, we exclude the 80 pixels at the top of the image and the 100 pixels at the bottom from the geometric difference. This prevents SDE artifacts in the sky and the hood of the ego car from contaminating the geometric difference.
The pixel-wise geometric difference is visualized in the third column of \autoref{fig:mg_showcase}. It can be observed that it is generally higher for samples that do not have a matching geometry or camera perspective. 

Based on a single target sample $i^{trg}$ and a set of candidate source samples $\mathcal{C}^{src}$, which are both sampled randomly, the source sample with the smallest geometric difference is selected for training
\begin{equation}
    j^{src} = \argmin_{c^{src} \in \mathcal{C}^{src}} G(i^{trg}, c^{src})\,.
\end{equation}

As the target sample is fixed during a matching step, it guides the selection towards the target distribution. The number of candidate samples $|C^{src}|$ balances between a good geometric match and a higher sampling diversity. A larger number of candidates results in a potentially better geometric match of the chosen sample, but it reduces the diversity of the chosen samples as it limits the sampling to the set of source samples that have a small geometric distance to the target domain in general.

This Matching Geometry Sampling allows our method to avoid the described issues of naive sampling and results in realistic DepthMix images, which are closer to the target distribution as can be seen in the last row of \autoref{fig:mg_showcase}.

\section{Experiment Setup}

\subsection{Datasets}
\label{sec:datasets}

\noindent\textbf{Cityscapes}: 
We mainly evaluate our method on the City\-scapes dataset \citep{cordts2016cityscapes}, which consists of 2975 training and 500 validation images with semantic segmentation labels from European street scenes. We downsample the images to $1024 \times 512$ pixels. Besides, random cropping to a size of $512 \times 512$ and random horizontal flipping are used during the training. Importantly, Cityscapes provides 20 unlabeled frames before and 10 after the labeled image, which are used for our SDE training. During the semi-supervised segmentation, only the 2975 images of the core dataset are used. If not stated otherwise, the same processing steps are applied to the following datasets as well.

\noindent\textbf{CamVid}:
The CamVid dataset \citep{brostow2009semantic} contains 367 training, 101 validation, and 233 test images with dense semantic segmentation labels for 11 classes from street scenes in Cambridge. To ensure a similar feature resolution as for Cityscapes, we upsample the CamVid images from $480 \times 360$ to $672 \times 512$ pixels and randomly crop them to a size of $512 \times 512$ pixels.

\noindent\textbf{GTA5}:
The GTA5 dataset \citep{richter2016playing} originates from a computer game, which enabled time-efficient semi\hyp{}automatic semantic segmentation annotation. It contains about 25k training images labeled using the same 19 classes as City\-scapes. The SDE is trained on another part of the dataset \citep{richter2017playing_sequences}, which provides image sequences.

\noindent\textbf{Synthia}:
The Synthia dataset \citep{germansellart2016large} provides synthetic images with automatically generated annotations from a simulated urban environment. For semantic segmentation, we use the  SYNTHIA-RAND-CITYSCAPES subset, which contains 9,400 samples labeled with 16 semantic classes common with Cityscapes. Following the standard protocol for domain adaptation, we train our method for the 16 semantic classes that are common with Cityscapes and evaluate on 13 of them. The SDE is trained on the SYNTHIA-SEQS video sequence subset.

\subsection{Network Architecture}
\label{sec:network_architecture}

Our network consists of a shared ResNet101 \citep{he2016deep} encoder with output stride 16, a decoder for segmentation, and a decoder for SDE. The decoder consists of an ASPP \citep{chen2017deeplab} block with dilation rates of 6, 12, and 18 to aggregate features from multiple scales and another four upsampling blocks with skip connections \citep{ronneberger2015u}. For SDE, the upsampling blocks have a disparity side output at the respective scale. For effective multi-task learning, we additionally follow PAD-Net \citep{xu2018pad} and deploy an attention-guided distillation module after the third decoder block. It serves the purpose of exchanging useful features between segmentation and depth estimation.
The design of the network architecture was chosen to facilitate both transfer and multi-task learning. To enable effective transfer learning, the task decoder branches have the same architecture and combine elements from typical semantic segmentation architectures such as the ASPP \citep{chen2017deeplab} as well as the commonly used U-Net decoder structure \citep{ronneberger2015u} for depth estimation. This allows for pretraining the segmentation decoder branch with SDE and repurposing it for semantic segmentation afterward.
For the pose estimation network, we use the same design as in \citep{godard2019digging}.
For the SDE network on the source domains, we use an output stride of 32 and a reduced number of decoder channels in order to improve convergence.

\subsection{Training}
\label{sec:training}

For the SDE pretraining, the depth and pose network are trained using the Adam optimizer, a batch size of 4, and an initial learning rate of $1 \times 10^{-4}$, which is divided by 10 after 160k iterations. The SDE loss is calculated on four scales with three subsequent frames. During the first 300k iterations, only the depth decoder and the pose network are trained. Afterwards, the depth encoder is fine-tuned with an ImageNet feature distance $\lambda_F = 1 \times 10^{-2}$ for another 50k iterations. The encoder is initialized with ImageNet weights, either before depth pretraining or before semantic segmentation if depth pretraining is ablated. The \emph{baseline} is trained with the same hyperparameters but only with a cross-entropy loss on the labeled samples. Its encoder is initialized with ImageNet pretrained weights.

For the semi-supervised multi-task learning, we train the network using SGD with a learning rate of $1 \times 10^{-3}$ for the encoder and depth decoder, $1 \times 10^{-2}$ for the segmentation decoder, and $1 \times 10^{-6}$ for the pose network. The learning rate is reduced by 10 after 30k iterations and the network is trained for another 10k iterations. A momentum of 0.9, a weight decay of $5 \times 10^{-4}$, and a gradient norm clipping to 10 are used. The loss for segmentation and SDE are weighted equally. The mean teacher has $\alpha=0.99$ and within an iteration, the network is trained on a clean labeled and an augmented mixed batch with size 2, respectively. The latter uses DepthMix with $\epsilon = 0.03$, color jitter, and Gaussian blur. 
If only pseudo-labeling but no mixing is used in an experiment, color jitter and Gaussian blur are still applied to the augmented batch.

For SSDA, the same hyperparameters as in the SSL setting are used. For the Matching Geometry Sampling, the number of random source candidate samples is set to $5$: $|C^{src}| = 5$.

\subsection{Automatic Data Selection for Annotation}
\label{sec:implementation_data_selection}

For the automatic data selection, we use a slimmed network architecture for $f_{SIDE}$ with a ResNet50 backbone, reduced decoder channels, and BatchNorm \citep{ioffe2015batch} in the decoder for efficiency and faster convergence. The depth student network is trained with a berHu loss using Adam with a learning rate of $1 \times 10^{-4}$ and polynomial decay with exponent 0.9.
For calculating the depth feature diversity, we use the output of the second depth decoder block after SDE pretraining. It is downsampled by average pooling to a size of 8x4 pixels and the feature channels are normalized to zero-mean and unit-variance over the dataset. The student depth error is weighted by $\lambda_E = 1000$. The number of the selected samples ($\sum_{t'=1}^t n_{t'}$) is incrementally increased to 25, 50, 100, 200, 372, and 744.
For each subset, a student depth network is trained from scratch for 4k, 8k, 12k, 16k, and 20k iterations, respectively, to calculate the student depth error and select the samples for the next subset.
The quality of the selected subset with annotations $\mathcal{G}_A$ is evaluated for semantic segmentation using our default architecture and training hyperparameters.
For the entropy baseline, a semantic segmentation network is trained on $\mathcal{G}_A$ and the samples with the highest mean pixel-wise Shannon entropy of the semantic segmentation prediction are greedily chosen from $\mathcal{G}_U$ to extend $\mathcal{G}_A$. Apart from that, the entropy baseline uses the same hyperparameters as our method.


\section{Results}

\subsection{Automatic Data Selection for Annotation}
\label{sec:exp_data_selection}

\begin{table}[b]
\centering
\caption{Comparison of data selection methods (DS: diversity sampling based on depth features, US: uncertainty sampling based on depth student error). mIoU in \%, std. dev. over 3 seeds.}
\label{tab:label_selection}
\begin{tabular}{llll}
\hline
\# Labeled          & 1/30 (100) & 1/8 (372) & 1/4 (744) \\
\hline\hline
Random & 48.75	\spm{1.61} & 59.14 \spm{1.02}	&	63.46	\spm{0.38}\\
Entropy & 53.63	\spm{0.77}	& 63.51	\spm{0.68}	& 66.18	\spm{0.50}\\
Ours (US) & 51.75	\spm{1.12}	& 62.77	\spm{0.46}	& 66.76	\spm{0.45}\\
Ours (DS) & 53.00	\spm{0.51}	& 63.23	\spm{0.69}	& 66.37	\spm{0.20}\\
Ours (DS+US) & \textbf{54.37}	\spm{0.36} & \textbf{64.25}	\spm{0.18}	& \textbf{66.94}	\spm{0.59} \\
\hline
\end{tabular}
\vspace*{\floatsep}
\centering
\caption{Comparison of the class-wise IoU in \% of the data selection methods for 372 labeled samples. The color visualizes the IoU difference with respect to the baseline.}
\label{fig:label_selection_classwise}
\includegraphics[width=1.0\linewidth]{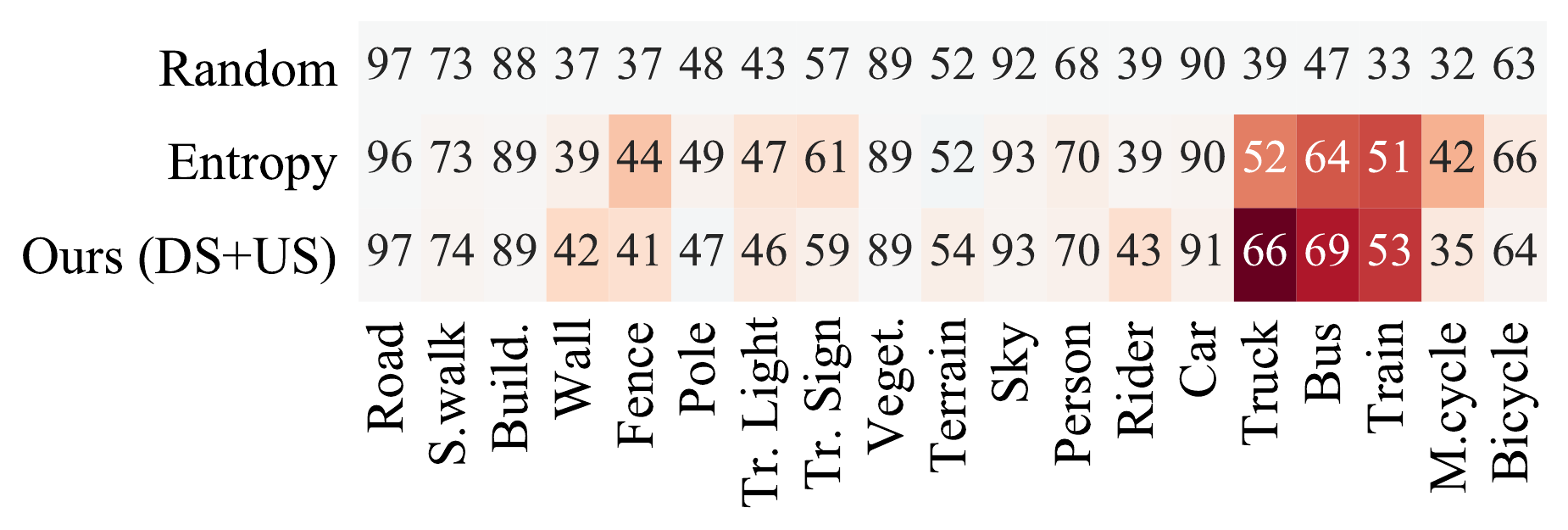}
\end{table}

\begin{figure*}
\centering
\includegraphics[width=0.9\linewidth]{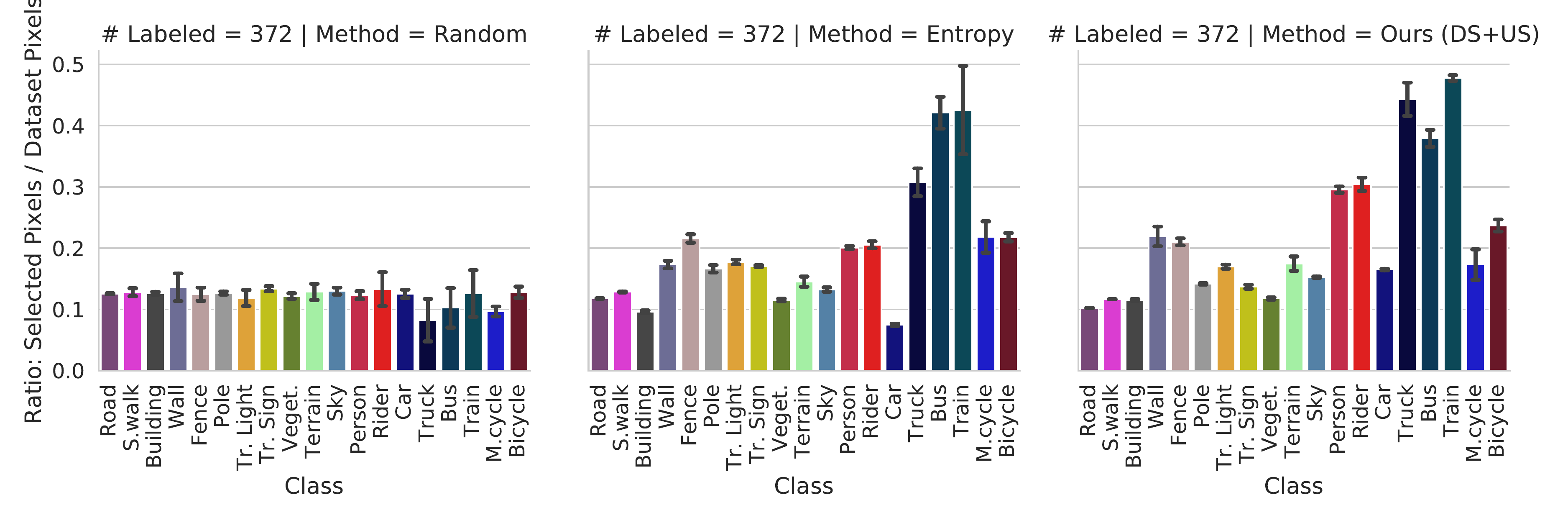}
\caption{Class frequency analysis of the data selection behavior. The ratio of selected pixels (372 samples) and dataset pixels (2975 samples) grouped by ground truth class for different data selection methods is shown. The values are averaged over 3 random seeds. The error bars indicate the standard deviation.}
\label{fig:label_selection_ratio}
\end{figure*}

First, we evaluate the proposed automatic data selection (see \autoref{sec:data_selection}) on the City\-scapes \citep{cordts2016cityscapes} dataset. \autoref{tab:label_selection} shows a comparison of our method with a baseline and a competing method for different numbers of selected labeled samples. The first baseline selects the labeled samples randomly, while the second, strong competitor uses active learning and iteratively chooses the samples with the highest segmentation entropy. In contrast to our method, this requires a human in the loop to create the semantic labels for iteratively selected images. \autoref{tab:label_selection} shows that our method with diversity sampling (DS) works better than with uncertainty sampling (US) for few labeled samples. We hypothesize that, for a small number of annotated samples, it is more important to better cover the underlying distribution with a diverse subset than just covering uncertain/difficult samples. For a larger subset, however, it makes sense to focus on the uncertain samples as the common cases are most likely already covered. Further, it can be seen that combining diversity sampling and uncertainty sampling (DS+US) performs better than using them individually showing that these criteria are complementary and cover two relevant aspects of selecting data for annotation. When comparing our method with both sampling criteria (DS+US) with the baselines ``Random" and ``Entropy", it can be seen that our method outperforms both comparison methods, demonstrating the effectiveness of ensuring diversity and exploiting difficult samples based on depth estimation. It also supports the assumption that depth estimation and semantic segmentation are correlated in terms of sample difficulty.
With 1/4 of the labeled samples, our method achieves 98.8~\% of the fully-supervised performance and with only 1/8 samples it still reaches 94.8~\%. Furthermore, the standard deviation of the achieved segmentation performance with our data selection is noticeably lower than for the random baseline when using few labeled samples, resulting in better reproducibility.

To better understand the underlying reasons for the improved performance, we analyze the class-wise IoU for 372 labeled samples in \autoref{fig:label_selection_classwise}. It shows that our automatic data selection significantly improves the performance of difficult classes with a low IoU of the random baseline such as wall, fence, truck, bus, and train. In comparison to the strong active learning entropy baseline, our method achieves even better performance for the classes wall, rider, truck, and bus.

In order to investigate possible reasons for the improved performance of the automatic data selection, we visualize the ratio of the automatically selected pixels and total dataset pixels grouped by the ground truth class for 372 selected samples in \autoref{fig:label_selection_ratio}. As expected, the ratio is about 0.125 for most of the classes when selecting 1/8 of the samples randomly (\autoref{fig:label_selection_ratio} left).
For the entropy baseline and our method, it can be seen that a higher ratio of difficult/rare classes (e.g. truck, bus, and train) are sampled from the underlying training set, while a smaller ratio of common classes such as road and building are sampled. When comparing the class-wise IoU (\autoref{fig:label_selection_classwise}) and the ratio of selected pixels (\autoref{fig:label_selection_ratio}), it can be seen that the improvement for difficult classes is correlated with them being selected more frequently by the automatic data selection. Intuitively, more samples of rare and easy to confuse classes such as car, truck, bus, and train as well as wall and fence will help the classifier to distinguish them. When comparing the active learning entropy baseline to our method, \autoref{fig:label_selection_ratio} shows that our method selects a higher ratio of wall, person, rider, and truck, which directly connects to the higher class IoU for these classes as shown in \autoref{fig:label_selection_classwise}.
Please note that the class-statistics of \autoref{fig:label_selection_ratio} are not available to our method during the entire selection process. This demonstrates that our method is able to correctly estimate the utility of samples for subsequent semantic segmentation without knowing the ground truth labels during the selection.


\subsection{DepthMix Data Augmentation}
\label{sec:exp_depthmix}

\begin{table}
\centering
\caption{Comparison of different mixing strategies. mIoU in \%, standard deviation over 3 seeds.}
\label{tab:mix}
\setlength{\tabcolsep}{2.5pt}
\begin{tabular}{lllll}
 \hline
 & \multicolumn{2}{l}{372 Labels} & \multicolumn{2}{l}{2975 Labels} \\
\hline\hline
Baseline & 59.14 \spm{1.02}  & \blarrow  & 67.77 \spm{0.13}  & \blarrow\\
Pseudo-Labels &     62.39 \spm{0.86}  & +3.24     & --                             &       \\
ClassMix &     63.16	\spm{0.89}   & +4.02    &   69.60	\spm{0.32} & +1.83 \\
DepthMix &     64.14	\spm{1.34} & +5.00 & 69.83	\spm{0.36} & +2.06 \\
\hline
\end{tabular}
\vspace*{\floatsep}
\centering
\caption{Comparison of the class-wise IoU in \% of the different mixing strategies for 372 labeled samples. The color visualizes the IoU difference with respect to the baseline.}
\label{fig:mix_classwise}
\includegraphics[width=1.0\linewidth]{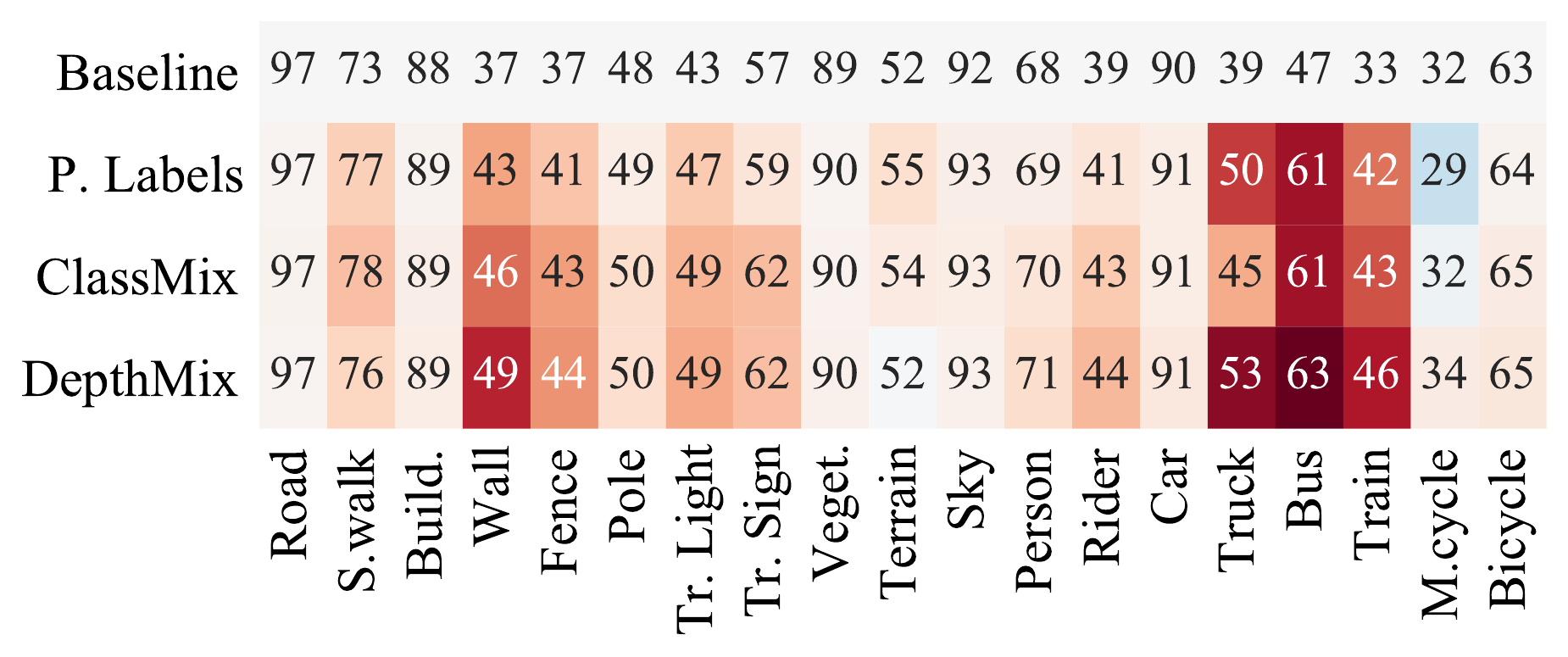}
\end{table}

Second, we study the proposed geometry-guided mixing strategy DepthMix (see \autoref{sec:methods_depthmix}). We evaluate the performance for the SSL setting with 372 of the labeled training samples (which corresponds to 1/8 of the labeled samples in Cityscapes) and the fully-supervised setting with 2975 samples. The subset of labeled samples is chosen randomly. \autoref{tab:mix} shows the mean and standard deviation of the mIoU in percent over three random seeds. Additionally, the improvement in percentage points of the analyzed components over the baseline, which only uses a cross-entropy loss on labeled samples, is shown.
In accordance with the literature on semi-supervised mixing \citep{french2019consistency, olsson2020classmix, sohn2020fixmatch}, we first add self-training with pseudo-labels from the mean teacher to the framework. As can be seen in \autoref{tab:mix}, this already significantly improves the performance in the SSL setting by +3.24 mIoU percentage points. Still, our proposed DepthMix module further increases the performance by another +1.76 (+2.06) percentage points for 372 (2975) labeled samples. 
Note that the high variance for few labeled samples is mostly due to the high influence of the randomly selected labeled subset. The chosen subset affects all configurations equally and the reported improvements are consistent for each subset.

When comparing DepthMix directly to the competitor ClassMix \citep{olsson2020classmix}, the performance of DepthMix is still +0.98 (+0.23) percentage point higher for 372 (2975) samples. This demonstrates the effectiveness of the geometry-aware mixing, which better handles occlusions as described in \autoref{sec:methods_depthmix}. The higher improvement of DepthMix for fewer labeled samples might be since the SDE for DepthMix can be trained on a large set of unlabeled samples, resulting in precise depth contours over the whole (un)labeled training set. ClassMix in contrast uses segmentation pseudo-labels for mixing, which were only supervised on the subset of labeled samples. Therefore, on the unlabeled samples, the mixing contours can be less accurate than for DepthMix.

Further, we analyze the class-wise IoU for 372 labeled samples as shown in \autoref{fig:mix_classwise}. Pseudo-labels generally improve the IoU through self-training. However, for the rare class motorcycle, the IoU decreases compared to the baseline. The reason for that is probably a pseudo-label drift of motorcycle towards the similar class bicycle during the self-training. Both mixing strategies mitigate the drift by a better generalization from labeled to unlabeled data through providing different contexts and occlusions during the training. The better generalization leads to less erroneous pseudo-labels and consequently to less drift. Additionally, this also results in a higher IoU for other difficult classes with a low baseline IoU such as sidewalk, wall, fence, traffic light, traffic sign, rider, truck, bus, and train. When comparing DepthMix and ClassMix, it can be seen that DepthMix improves over ClassMix for difficult classes with usually pronounced depth contours such as wall, traffic light, rider, bus, train, and motorcycle. However, there is a slight decrease in IoU for the classes sidewalk and terrain. These are classes, which can be easily confused with each other and with road. DepthMix might experience difficulties with these classes as there are usually no depth contours between them, which results in fewer mixing boundaries.

\begin{figure*}
    \centering
    a) \includegraphics[width=0.96\linewidth,valign=c]{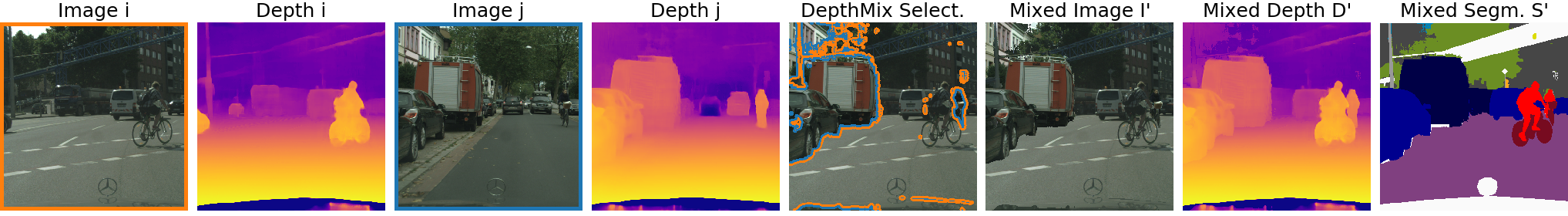}\\
    b) \includegraphics[width=0.96\linewidth,valign=c]{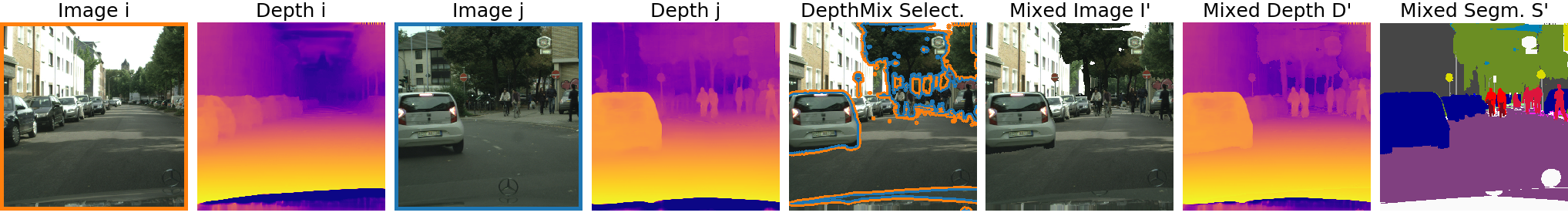}\\
    c) \includegraphics[width=0.96\linewidth,valign=c]{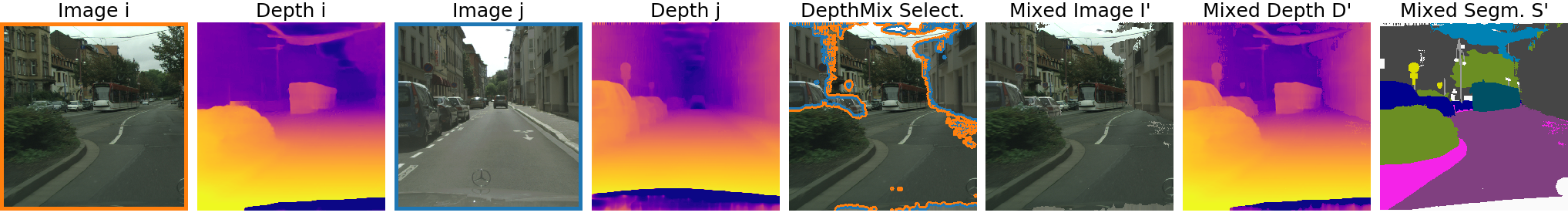}\\
    \vspace{0.2cm}
    \par\noindent\rule{\textwidth}{1.0pt}\\
    \vspace{0.2cm}
    d) \includegraphics[width=0.96\linewidth,valign=c]{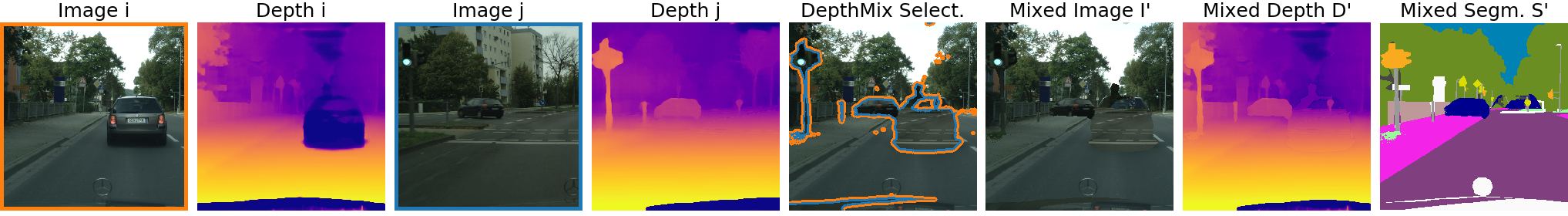}\\
    e) \includegraphics[width=0.96\linewidth,valign=c]{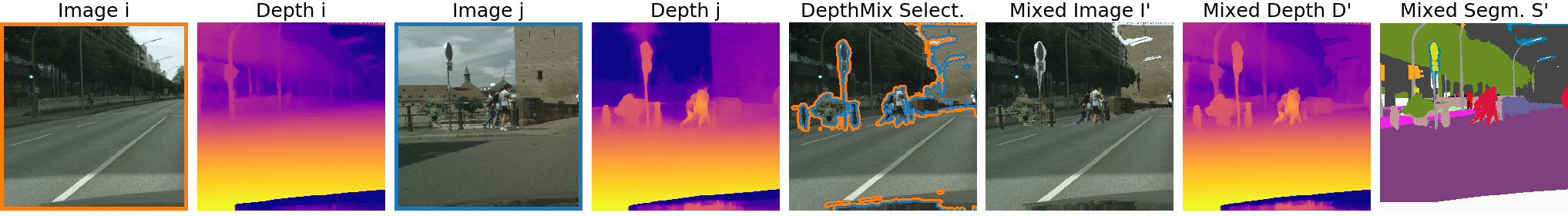}
    \caption{Examples of DepthMix applied to Cityscapes crops. From left to right, the source images with their SDE estimate, the mixed image $I'$ overlaid with the border of the mix mask $M$ in blue/orange depending on the adjacent source image (i - orange, j - blue), the mixed image without visual guidance $I'$, the mixed depth $D'$, and the mixed segmentation $S'$ are shown. For simplicity, the source segmentations for the mixed segmentation $S'$ originate from the ground truth labels. Rows a) -- c) demonstrate the strength of DepthMix to handle occlusions, while rows d) and e) show typical failure cases}
    \label{fig:depthmix_examples}
\end{figure*}

The effective occlusion handling of DepthMix can be seen in \autoref{fig:depthmix_examples}~a)~--~c) for samples from Cityscapes. It shows input images in orange and blue as well as their SDE used for mixing. The column ``DepthMix Select." visualizes from which input image the regions, chosen by DepthMix, originate. As can be seen in \autoref{fig:depthmix_examples}~a), DepthMix is able to handle occlusions at multiple levels. The biker from the blue image occludes buildings from the orange image, but the blue biker is itself also partly occluded by the closer biker from the orange image. Similar cases can be seen for trees, traffic signs, and cars in \autoref{fig:depthmix_examples}~~b) and c). The column ``Mixed Image $I'$" shows the resulting image without the selection overlay. It can be seen that due to the spatially accurate depth contours, the mixed images contain only minor mixing border artifacts and have a realistic appearance. The same is the case for the mixed segmentation as can be seen in the column ``Mixed Segm. $S'$"
    
However, there are also some cases in which DepthMix fails to correctly mix images according to their geometry. Examples of typical failure cases are shown in \autoref{fig:depthmix_examples}~d) and e). First, the SDE can be inaccurate for dynamic objects due to the violation of the static world assumption, which can cause an inaccurate structure within the mixed image. This is particularly the case if a car is driving in front of the ego car (\autoref{fig:depthmix_examples}~d)). However, this type of failure case is common in ClassMix and its frequency is greatly reduced with DepthMix. A remedy might be SDE extensions that incorporate the motion of dynamic objects \citep{casser2019depth, dai2020self, klingner2020self}. Second, in some cases, the SDE can be imprecise and the depth discontinuities do not appear at the same location as the class border. This can cause artifacts in the mixed image as well as in the mixed segmentation as can be seen for the sky within the building in \autoref{fig:depthmix_examples}~e). Note that the same can happen for ClassMix when the pseudo-labels, used for the mixing, do not have accurate segmentation borders.


\subsection{Transfer and Multi-Task Learning}
\label{sec:exp_mtl}

Third, we study the proposed transfer and multi-task learning of semantic segmentation and the auxiliary task self-supervised depth estimation. 
For 372 (2975) samples, SDE transfer learning of the encoder and decoder (with previous ImageNet pretraining of the encoder) improves performance by +1.31 (+1.23) percentage points mIoU over the baseline with only ImageNet pretraining of the encoder. This demonstrates the usefulness of the features learned by SDE for semantic segmentation, both in the semi- and fully-supervised case. Additional regularization of the encoder with an ImageNet feature distance loss during SDE pretraining improves the performance by another +0.35 (+0.48) percentage points. Furthermore, multi-task learning in addition to transfer learning results in a performance increase of +0.45 (+0.29) percentage points.

\begin{table}
\centering
\caption{Comparison of SDE feature transfer methods (F: ImageNet feature distance loss). mIoU in \%, standard deviation over 3 seeds.}
\label{tab:ablation_mtl}
\setlength{\tabcolsep}{2.5pt}
\begin{tabular}{ccllll}
 \hline
Aux. SDE & F & \multicolumn{2}{l}{372 Labels} & \multicolumn{2}{l}{2975 Labels} \\
\hline\hline
  &     &  59.14 \spm{1.02}  & \blarrow  & 67.77 \spm{0.13}  & \blarrow\\
Transfer &     &  60.46 \spm{0.64}  & +1.31     & 69.00 \spm{0.70}  & +1.23 \\
Transfer & \checkmark & 60.80 \spm{0.69}  & +1.66     & 69.47 \spm{0.38}  & +1.71 \\
Multi-Task & \checkmark & 61.25 \spm{0.55}  & +2.10     & 69.76 \spm{0.39}  & +1.99 \\
\hline
\end{tabular}
\vspace*{\floatsep}
\centering
\caption{Comparison of the class-wise IoU in \% of SDE feature transfer methods for 372 labeled samples (F: ImageNet feature distance loss). The color visualizes the IoU difference with respect to the baseline.}
\label{fig:mtl_classwise}
\includegraphics[width=1.0\linewidth]{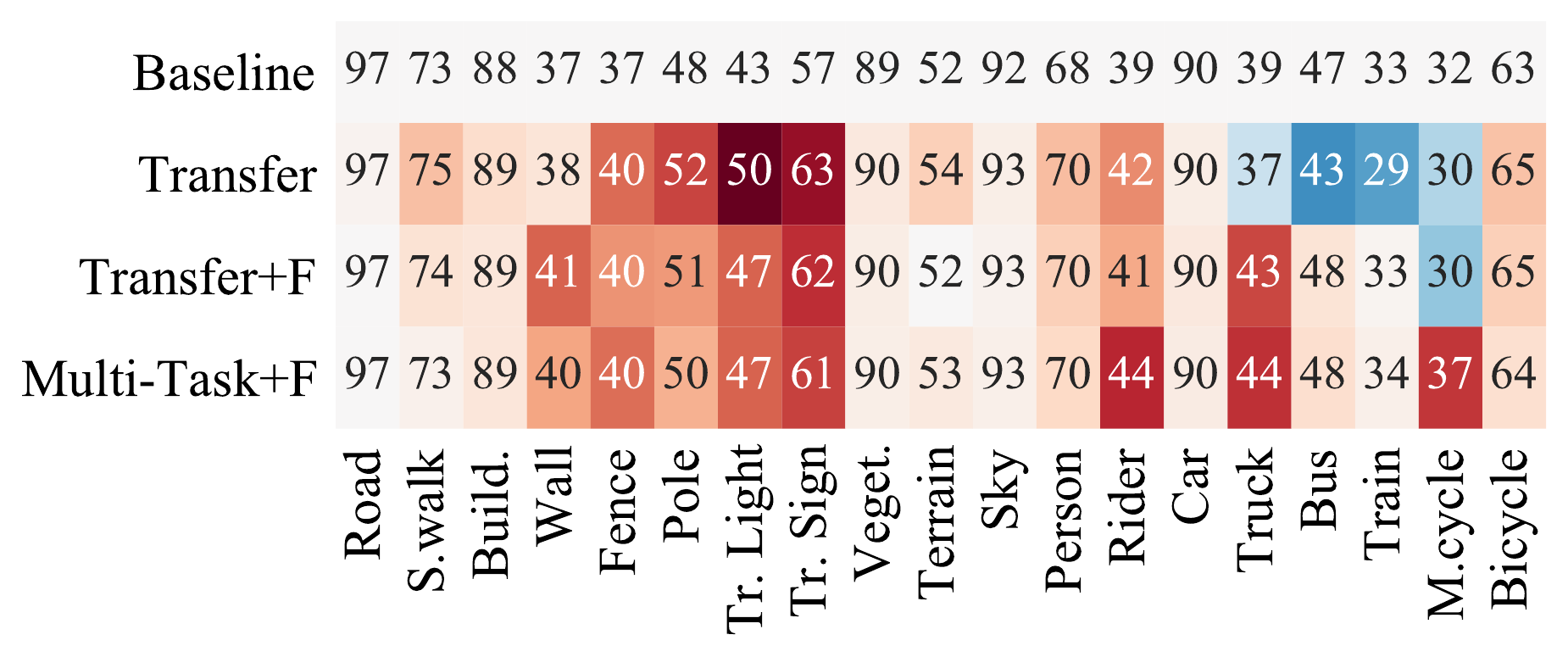}
\end{table}

The class-wise analysis for 372 labeled samples (see \autoref{fig:mtl_classwise}) shows that SDE transfer learning without ImageNet Feature distance loss significantly improves the performance of classes, where segmentation border coincides with depth discontinuities such as fence, pole, traffic light, and traffic sign. This is possibly due to their characteristic depth profile learned during SDE. For example, a good depth estimation performance requires correctly segmenting poles or traffic signs as missing them can cause large depth errors.
However, there is a performance drop for classes that have slight semantic differences such as truck, bus, train, and motorcycle. We hypothesize that the SDE pretraining causes forgetting important semantic features from the ImageNet pretraining that are relevant for semantic segmentation but not for SDE. For example, for SDE it is not relevant if an object is a bus or a train but for semantic segmentation it is. Adding the ImageNet feature distance loss to the SDE pretraining in order to avoid forgetting these semantic features, prevents the performance drop for truck, bus, and train. The additional multi-task learning further improves the performance for the small difficult classes rider and motorcycle.


\subsection{Combined Framework for SSL}
\label{sec:exp_combined}

Next, we combine the three contributions multi-task learning, DepthMix, and automatic data selection for annotation into a unified semi-supervised semantic segmentation framework. The first part of \autoref{tab:combined_semi_supervised} summarizes the performance of these components from the previous sections for a better comparison. The component with the most improvement is the automatic data selection for annotation with diversity and uncertainty sampling with +5.11 mIoU percentage points for 372 labeled samples. However, it is not applicable to the full dataset as there is no need for sample selection -- all samples are used. The second-most effective component is DepthMix with pseudo-labeling, which also has a pronounced mIoU improvement of +5.00 (+2.06) for 372 (2975) samples. The smallest but still significant improvement comes from multi-task learning with +2.00 (+1.99) percentage points. The direct comparison of the class-wise IoU for 372 labeled samples in \autoref{fig:classwise_improvement} shows that data selection mostly improves the performance of difficult classes with a low baseline IoU (e.g. wall, fence, truck, bus, and train), SDE multi-task learning of classes with surrounding depth discontinuities (e.g. fence, pole, traffic light, traffic sign, and rider), and DepthMix of both.

\begin{table}
\centering
\caption{Comparison of the combinations of the proposed framework components (S: data selection, DX: DepthMix, MTL: SDE multi-task learning). mIoU in \%, standard deviation over 3 seeds.}
\label{tab:combined_semi_supervised}
\setlength{\tabcolsep}{4pt}
\begin{tabular}{cccllll}
 \hline
S   & DX  & MTL  & \multicolumn{2}{l}{372 Labels} & \multicolumn{2}{l}{2975 Labels} \\
\hline\hline
    &     &      & 59.14 \spm{1.02}  & \blarrow  & 67.77 \spm{0.13}  & \blarrow\\
    &     & \cm  & 61.25 \spm{0.55}  & +2.10     & 69.76 \spm{0.39}  & +1.99 \\
    & \cm &      & 64.14 \spm{1.34}  & +5.00     & 69.83 \spm{0.36}  & +2.06 \\
\cm &     &      & 64.25 \spm{0.18}  & +5.11     & --                & \\
\hline
\cm &     & \cm  & 65.35 \spm{0.10}  & +6.21     & --                &  \\
\cm & \cm &      & 66.48 \spm{0.27}  & +7.34     & --                &  \\
    & \cm & \cm  & 66.66 \spm{1.05}  & +7.52     & 71.16 \spm{0.16}  & +3.40 \\
\hline
\cm & \cm & \cm  & 68.01 \spm{0.83}   & +8.87    & --                & \\
\hline
\end{tabular}
\vspace*{\floatsep}
\centering
\caption{Comparison of the class-wise IoU in \% of the combinations of the proposed framework components for 372 labeled samples (see \autoref{tab:combined_semi_supervised} for abbreviations). The color visualizes the IoU difference with respect to the baseline.}
\label{fig:classwise_improvement}
\includegraphics[width=\linewidth]{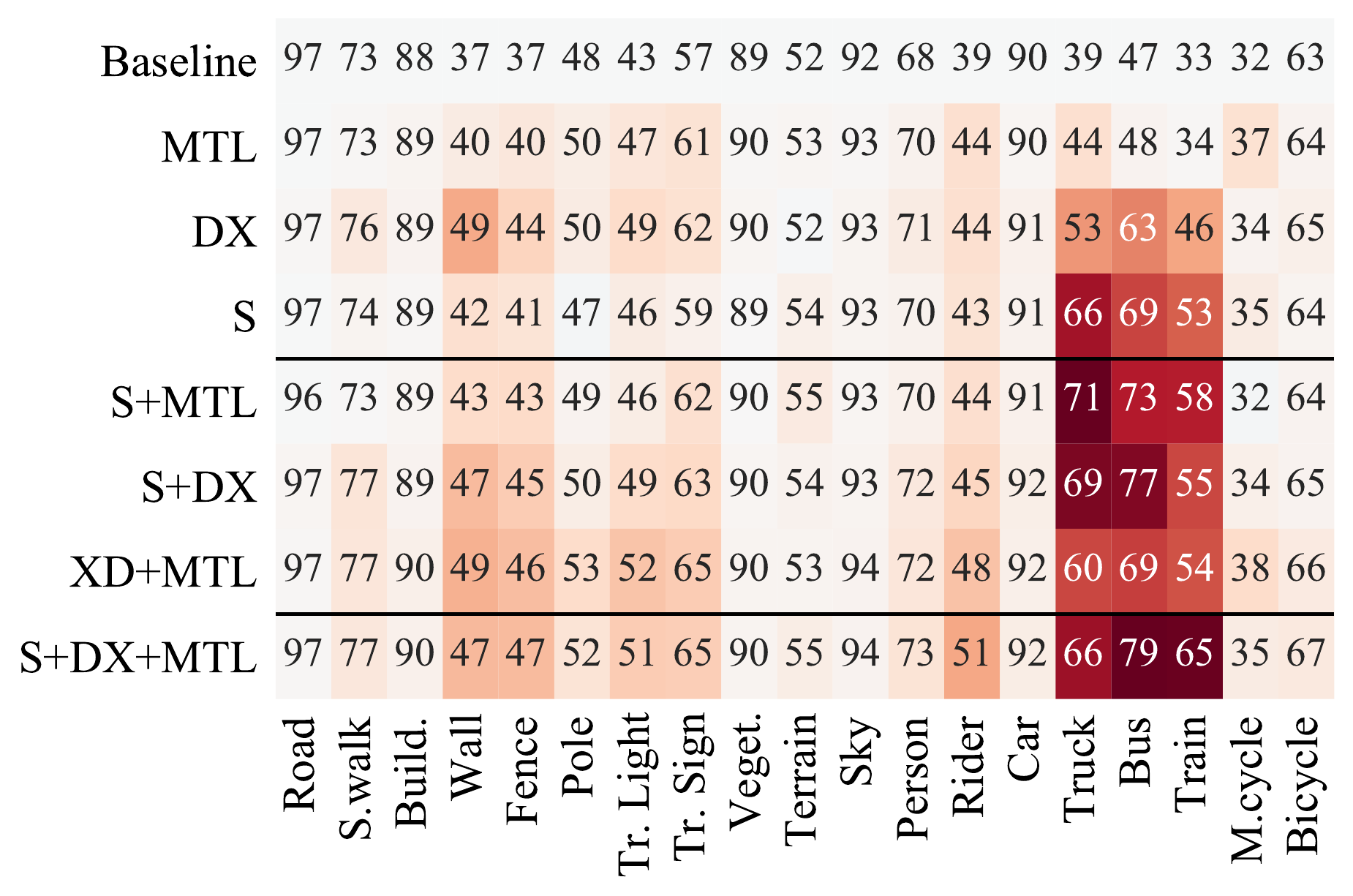}
\end{table}

Considering that the three contributions follow different approaches and improve the performance of a different subset of classes, we further study the combination of our contributions as shown in the second part of \autoref{tab:combined_semi_supervised} and \autoref{fig:classwise_improvement}. The improvement over the baseline performance is +6.21 when combining multi-task learning with data selection, +7.34 when combining DepthMix and data selection, and +7.52 (+3.40) when combining multi-task learning and DepthMix for 372 (2975) samples. In all cases, the combination is better than every single component. The class-wise analysis for 372 labeled samples in \autoref{fig:classwise_improvement} reveals that the class performance of the combination usually is the highest class performance of the components. As the components perform well on different classes, this already attributes to the improved performance of the combinations. Moreover, there are some classes such as fence, traffic sign, rider, truck, bus, and train, where the performance of the combination is even higher than its best component. This might be due to self-reinforcing effects. For example, the improved segmentation detail at depth contours from multi-task learning is propagated into DepthMix and results in even better pseudo-label supervision for mixed samples. The last row of \autoref{tab:combined_semi_supervised} shows the combination of all three contributions. With an improvement of +8.87 percentage points for 372 labeled samples, it achieves the best results so far. It combines the strength of our three contributions and significantly improves the performance for classes with depth discontinuities and for difficult classes. The most improvement is achieved for truck, bus, and train, where the mIoU is more than 50\% better than the baseline.


\subsection{Comparison with State-of-the-Art SSL Methods}
\label{sec:sota_comparison}

\begin{table*}
\begin{minipage}{\textwidth}
\centering
\caption{Comparison with state-of-the-art SSL semantic segmentation methods on the Cityscapes validation set (mIoU in \%, standard deviation over 3 random seeds). The best results are shown in bold font and the second-best results are underlined.}
\label{tab:comp}
\setlength{\tabcolsep}{3pt}
\begin{tabular}{|l|ll|ll|ll|ll|}
\hline
Labeled Samples & 1/30 (100)    &        & 1/8 (372)     &       & 1/4 (744)     &       & Full (2975) &       \\\hline\hline
Baseline \citep{hung2018adversarial}        & --             &        & 55.50 & \blarrow      & 59.90 & \blarrow      & 66.40     & \blarrow      \\
Adversarial \citep{hung2018adversarial}      & --       &        & 58.80 & +3.30 & 62.30 & +2.40 & --           &       \\\hline\hline
Baseline \citep{mittal2019semi}       & --             &        & 56.20 & \blarrow      & 60.20 & \blarrow      & 66.00     &      \\
s4GAN \citep{mittal2019semi}         & --       &        & 59.30 & +3.10 & 61.90 & +1.70 & 65.80     & --0.20 \\\hline\hline
Baseline \citep{feng2020semi}        & 45.50       & \blarrow          & 56.70 & \blarrow      & 61.10 & \blarrow      & 66.90     &      \\
DST--CBC \citep{feng2020semi}    & 48.70   & +3.20  & 60.50 & +3.80 & 64.40 & +3.30 & --           &       \\\hline\hline
Baseline \citep{fenga2020dmt} & 49.54 & \blarrow & 59.65 & \blarrow & -- &  & 68.16           &       \\
DMT \citep{fenga2020dmt}     & 54.80 & +5.26 & 63.03 & +3.38 & -- &  & --           &       \\\hline\hline
Baseline \citep{french2019consistency}        & 44.41 \scriptsize{$\pm1.11$}       & \blarrow          & 55.25 \scriptsize{$\pm0.66$} & \blarrow      & 60.57 \scriptsize{$\pm1.13$} & \blarrow      & 67.53 \scriptsize{$\pm0.35$}     & \blarrow      \\
CutMix \citep{french2019consistency}        & 51.20 \scriptsize{$\pm2.29$} & +6.79  & 60.34 \scriptsize{$\pm1.24$} & +5.09 & 63.87 \scriptsize{$\pm0.71$} & +3.30 & 67.68 \scriptsize{$\pm0.37$}     & +0.15 \\\hline\hline
Baseline \citep{mendel2020semi}        & --      &    & 55.96 \scriptsize{$\pm0.86$} & \blarrow      & 60.54 \scriptsize{$\pm0.85$} & \blarrow      & --  &      \\
ECS \citep{mendel2020semi}        & -- &  & 60.26 \scriptsize{$\pm0.84$} & +4.30 & 63.77 \scriptsize{$\pm0.65$} & +3.23 & --           &       \\\hline\hline
Baseline \citep{olsson2020classmix}        & 43.84 \scriptsize{$\pm0.71$}      & \blarrow       & 54.84 \scriptsize{$\pm1.14$} & \blarrow      & 60.08 \scriptsize{$\pm0.62$} & \blarrow      & 66.19 \scriptsize{$\pm0.11$}    &      \\
ClassMix \citep{olsson2020classmix}        & 54.07 \scriptsize{$\pm1.61$} & \underline{+10.23} & 61.35 \scriptsize{$\pm0.62$} & +6.51 & 63.63 \scriptsize{$\pm0.33$} & +3.55 & --           &       \\\hline\hline
ATSO \citep{huo2021atso}\footnote{ATSO does not provide baseline results.}        & 53.1 &  & 61.8 &  & 63.2 &  & --           &       \\\hline\hline
Baseline & 48.75	\spm{1.61} & \blarrow &	59.14	\spm{1.02}	& \blarrow &	63.46	\spm{0.38}	& \blarrow &	67.77	\spm{0.13} & \blarrow \\
ClassMix \citep{olsson2020classmix}\footnote{Results of the reimplementation in our experiment setting.} & 56.82	\spm{1.65} &	+8.07 &	63.86	\spm{0.41} &	+4.72 &	65.57	\spm{0.71} &	+2.11 &	--		& \\
ClassMix \citep{olsson2020classmix} (+Video) & 56.79	\spm{1.98} & +8.04 & 63.22 \spm{0.84} & +4.08 &	65.72 \spm{0.18} & +2.26 & \underline{68.23}	\spm{0.70}		& +0.46 \\
Ours w/o Data Selection	& \underline{58.40}	\spm{1.36} &	+9.65 &	\underline{66.66}	\spm{1.05} &	\underline{+7.52} &	\underline{68.43}	\spm{0.06} &	\underline{+4.98} &	\textbf{71.16}	\spm{0.16} &	\textbf{+3.40} \\
Ours & \textbf{62.09}	\spm{0.39} &	\textbf{+13.34} &	\textbf{68.01}	\spm{0.83} &	\textbf{+8.87} &	\textbf{69.38}	\spm{0.33} &	\textbf{+5.92} & -- & \\
\hline
\end{tabular}
\end{minipage}
\end{table*}

\begin{figure*}
    \centering
    \includegraphics[width=\linewidth]{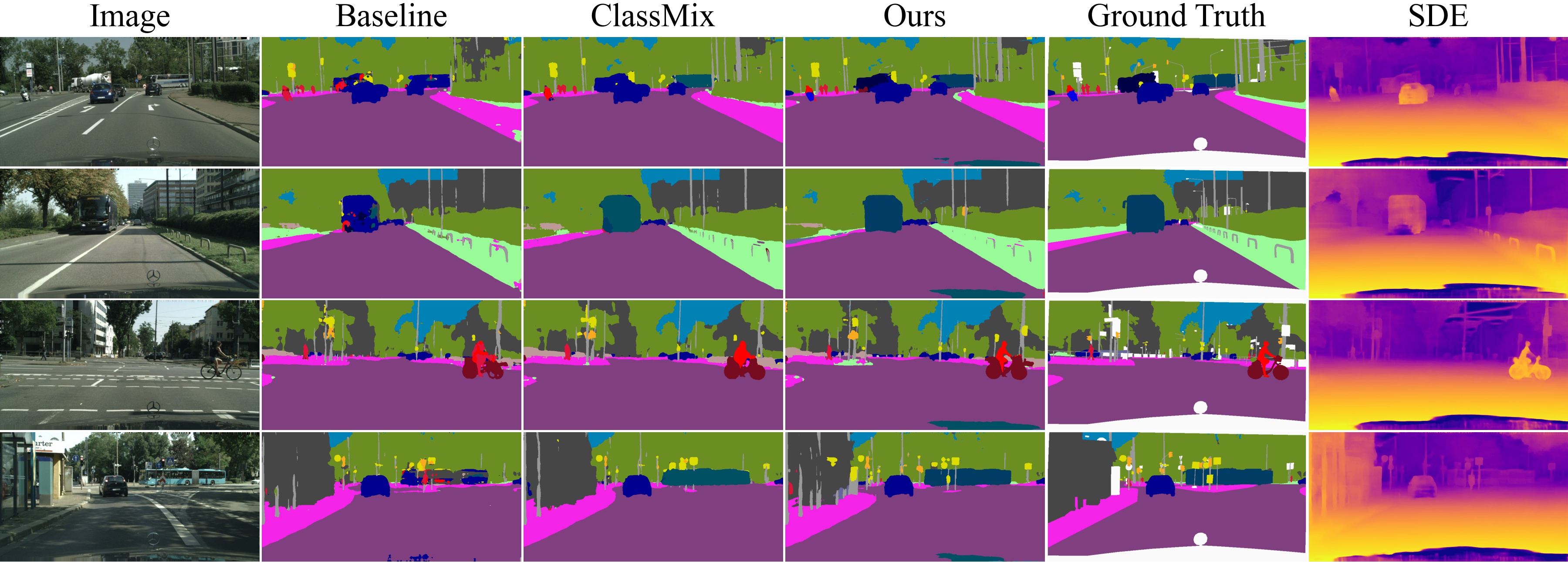}
    \caption{Example semantic segmentations and self-supervised depth estimates of our method for 100 labeled samples in comparison with ClassMix \citep{olsson2020classmix} and the baseline.}
    \label{fig:examples}
\end{figure*}

Next, we compare our approach with several state-of-the-art SSL approaches. The results are summarized in \autoref{tab:comp}. The performance (mIoU in \%) of the SSL methods and their baselines (which use the same backbone network but are only trained on the labeled dataset) are shown over a different number of labeled samples. As the performance of the baselines differs, there are columns showing the absolute improvement for better comparability. As our baseline utilizes a more capable network architecture due to the U-Net decoder with ASPP as opposed to a DeepLabv2 decoder used by most previous works, we also reimplemented the state-of-the-art method, ClassMix \citep{olsson2020classmix} with our network architecture and training parameters to ensure a direct comparison.

As shown in \autoref{tab:comp}, our method (without data selection) outperforms all other approaches on each labeled subset size for both the absolute performance as well as the improvement to the baseline. 
The only exception is the absolute improvement of the original results of ClassMix for 100 labeled samples. However, if we consider ClassMix trained in our setting, our method outperforms it also in this case. This can be explained by the considerably higher baseline performance in our setting, which increases the difficulty to achieve a high improvement. 
Adding data selection even further increases the performance by a significant margin, so that our method, trained with only 1/8 of the labels, even slightly outperforms the fully-supervised baseline.

To identify whether the improvement originates from access to more unlabeled data or from the effectiveness of our approach, we compare it to another baseline ``ClassMix (+Video)". More specifically, we also provide all unlabeled image sequences to ClassMix and see how much it can benefit from this additional amount of unlabeled data. Experimental results show no significant difference. This is probably due to the high correlation between the Cityscapes image dataset and the video dataset (the images are the 20th frames of the video clips).

The adequacy of our approach is also reflected in the example predictions in \autoref{fig:examples}. We can observe that the contours of classes are more precise. This is particularly the case for classes, which are surrounded by depth discontinuities such as poles, traffic signs, rider, or person. Moreover, difficult objects such as bus, train, rider, or truck can be better distinguished. As discussed in \autoref{sec:exp_combined}, this observation is also quantitatively confirmed by the class-wise IoU improvement shown in \autoref{fig:classwise_improvement}.
On the downside, SDE sometimes fails for cars driving directly in front of the camera (see 7th row in \autoref{fig:examples}) and violating the reconstruction assumptions. Those cars are observed at the same location across the image sequence and can not be correctly reconstructed during SDE training, even with correct depth and pose estimates. However, the network-internal differentiation between moving and non-moving cars does not hinder the transfer of SDE-learned features to semantic segmentation but can cause problems with DepthMix (see \autoref{sec:exp_depthmix}).

\subsection{Learning SDE and Semantic Segmentation on Different Datasets}
\label{sec:different_datasets}

\begin{table}
\centering
\caption{Semantic segmentation performance on the CamVid test set with SDE trained on Cityscapes sequences (mIoU in \%, standard deviation over 3 random seeds).}
\label{tab:camvid}
\setlength{\tabcolsep}{3pt}
\begin{tabular}{lllllll}
\hline
\# Labeled & 50    &        & 100     &       & 367 (Full) &       \\\hline\hline
Baseline & 59.2	\spm{1.8} & \blarrow &		63.1	\spm{0.6} & \blarrow &		68.2	\spm{0.1} & \blarrow \\
ClassMix & 65.9	\spm{0.3} &	+6.7 &	67.5	\spm{1.0} &	+4.4 &	-	&  \\
Ours w/o S & 66.8	\spm{1.2} &	+7.6 &	68.9	\spm{0.6} &	+5.8 &	\textbf{71.5}	\spm{0.2} &	\textbf{+3.3} \\
Ours & \textbf{68.2}	\spm{0.4} &	\textbf{+9.0} &	\textbf{69.6}	\spm{0.6} &	\textbf{+6.5} &	- & \\
\hline
\end{tabular}
\end{table}

In this section, we show that the unlabeled image sequences and the labeled segmentations can also originate from different datasets within similar visual domains. For that purpose, we train the SDE on Cityscapes sequences and learn the semi-supervised semantic segmentation on the CamVid dataset \citep{brostow2009semantic}. As we assume in this scenario that there are no image sequences available for SDE training on CamVid, we only apply transfer learning but no multi-task learning. 

\autoref{tab:camvid} shows that the results on CamVid are similar to our main results on Cityscapes. For 50/100/367 labeled training samples, our method improves the mIoU by +9.0/\allowbreak+6.5/+3.3 percentage points. In the end, our proposed method significantly outperforms ClassMix \citep{olsson2020classmix} by +2.3 percentage points for 50 labeled samples and +2.1 percentage points for 100 labeled samples.

\subsection{Component Study for SSDA}
\label{sec:exp_ssda}

\begin{table}[b]
\centering
\caption{Comparison of the previous framework components in a SSDA setting (SD: additional source domain data, S: data selection, DX: DepthMix, MTL: SDE multi-task learning).}
\label{tab:ssda_naive}
\setlength{\tabcolsep}{2.5pt}
\begin{tabular}{cccccccc}
\hline
SD & S   & DX & MTL & 100 Trg. Labels & 500 Trg. Labels \\
\hline\hline
   &     &   &      & 48.75 \spm{1.62} & 61.66 \spm{0.90} \\
   & \cm &\cm& \cm  & 62.09 \spm{0.39} & 67.75 \spm{0.10} \\
\hline
\cm&     &   &      & 53.83	\spm{1.09} & 60.99 \spm{1.04} \\
\cm&     &   & \cm  & 56.20	\spm{0.92} & 62.46 \spm{1.04} \\
\cm&     &\cm&      & 60.05	\spm{1.91} & 66.19 \spm{0.80} \\
\cm& \cm &   &      & 54.92	\spm{0.68} & 61.97 \spm{0.74} \\
\cm& \cm &\cm&\cm   & 64.54	\spm{0.12} & 68.63 \spm{0.34} \\
\hline
\end{tabular}
\vspace*{\floatsep}
\centering
\caption{Comparison of the class-wise IoU in \% of the previous framework components in a SSDA setting for 100 labeled target samples (see \autoref{tab:ssda_naive} for abbreviations). The color visualizes the IoU difference with respect to SD.}
\label{fig:ssda_naive_classwise}
\includegraphics[width=1.0\linewidth]{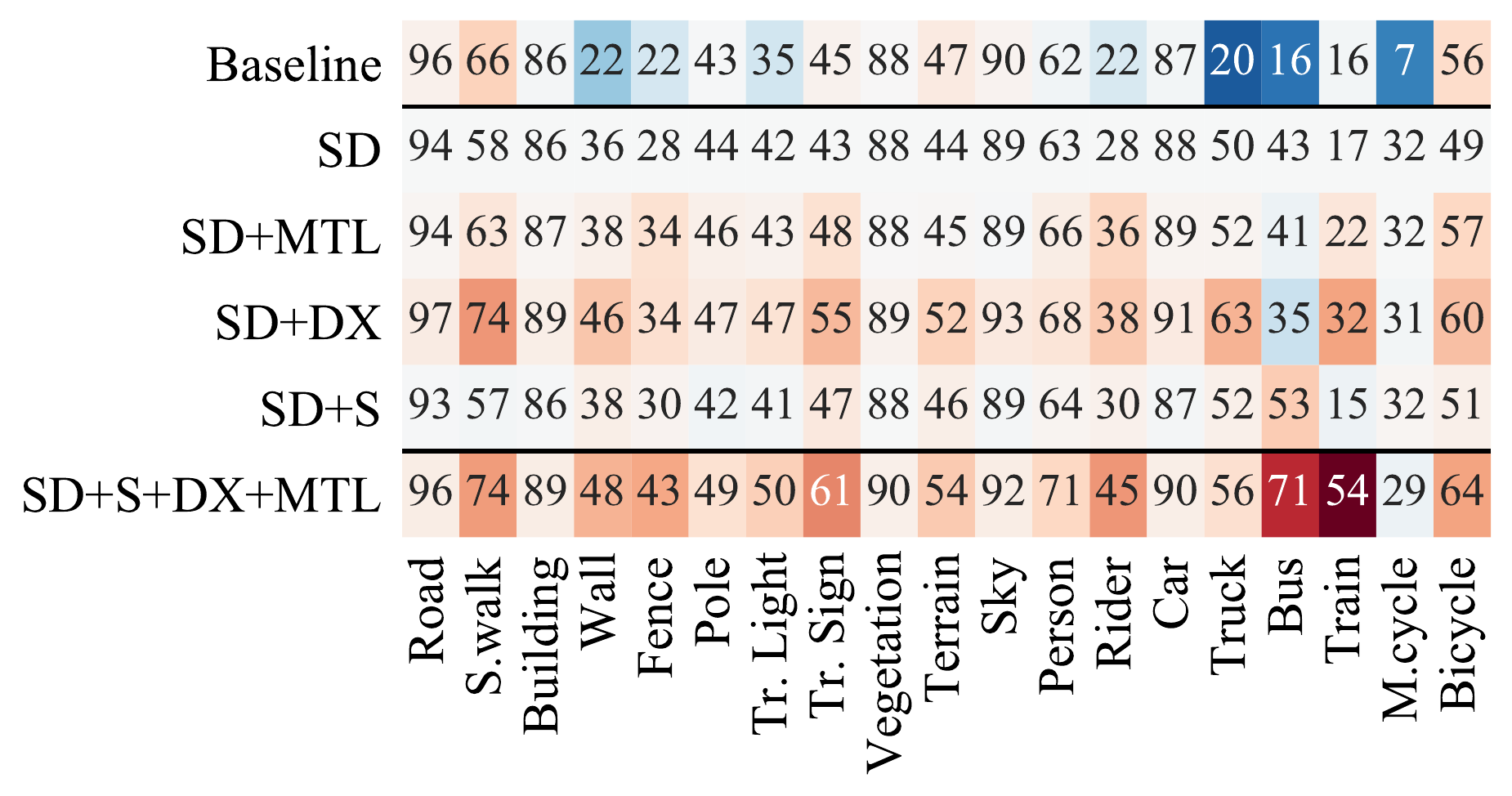}
\end{table}

We study the components of the SSDA framework described in \autoref{sec:methods_ssda} on the commonly used benchmark GTA5 $\rightarrow$ Cityscapes, where the synthetic source training samples originate from the GTA5 dataset \citep{richter2016playing} and the real target training samples are obtained from Cityscapes \citep{cordts2016cityscapes}. After the training, the network is evaluated on the target validation samples from the Cityscapes validation set. First, we analyze our contributions from SSL in an SSDA setting by naively adding the additional source samples to the training according to \autoref{eq:clean_source_loss}. The remaining framework is the same as in the previous experiments.

The first part of \autoref{tab:ssda_naive} shows the results using the SSL framework without source domain supervision, while the second part shows the results for the framework with additional semantic segmentation supervision from the source domain according to \autoref{eq:clean_source_loss}.

For 100 labeled samples from the target domain, \autoref{tab:ssda_naive} shows that additional source domain supervision improves the performance of the baseline by +5.08 percentage points. 
As can be seen in \autoref{fig:ssda_naive_classwise}, this is mainly due to improvements for classes with a low baseline performance such as wall, fence, traffic light, rider, truck, bus, and motorcycle. However, additional source domain supervision deteriorates the performance for the classes sidewalk, terrain, and bicycle, which are easy to confuse and have a considerable domain gap. 
When applying our proposed methods from SSL, they also lead to an improved performance in the SSDA setting as shown in the second part of \autoref{tab:ssda_naive}. 
For multi-task learning, the gain is +2.37 percentage points with the same performance pattern of the class-wise IoU. For DepthMix, the improvement is +6.22, while it also effectively counters the performance drop (from Baseline to SD) for the classes road, sidewalk, terrain, and bicycle (see \autoref{fig:ssda_naive_classwise}). For automatic data selection, the improvement by additional source data is +1.09.
When combining the three contributions, the performance gain over the baseline with source supervision is +10.71. This is +2.45 percentage points better than our method for SSL.

For 500 labeled samples from the target domain, additional source domain supervision decreases the performance for the baseline by -0.67 percentage points (see \autoref{tab:ssda_naive}). This shows that additional source supervision is not helpful in this case, probably, because there is already decent supervision on the target domain and naively adding the source domain loss cannot close the domain gap. But also in this setting, multi-task learning / DepthMix / data selection can still improve the performance by +1.47 / +5.2 / +0.98 over the baseline with source supervision. When being combined, their performance gain is +7.64. This is +0.88 percentage point better than our method for SSL.

Next, we analyze our contributions tailored to overcome the domain gap of SSDA: Cross-Domain DepthMix (see \autoref{sec:methods_CDM}) and Matching Geometry Sampling (see \autoref{sec:methods_mg}). 
\autoref{tab:ssda_advanced} shows that both Cross-Domain DepthMix (CDM) and Target-Domain DepthMix (TDM) significantly outperform the baseline. As shown in \autoref{fig:ssda_advanced_classwise}, this is due to an improved performance for difficult classes such as sidewalk, wall, traffic sign, terrain, rider, truck, train, and motorcycle. Through DepthMix presenting these objects with different backgrounds and occlusions, the network learns to generalize better within the target domain (for TDM) or across domains (CDM).
When comparing the performance of CDM and TDM (see \autoref{tab:ssda_advanced}), it can be seen that CDM works better for 100 labeled target samples and TDM works better for 500. 
On the one side, CDM can exploit the labeled source data to propagate its knowledge to the target data through mixing. This is especially useful if there are only a few labeled target samples available and most supervision comes from the source domain.
On the other side, TDM can use the already labeled target samples to propagate their knowledge to the unlabeled target through mixing, without being impeded by a domain gap. This is most effective when there are sufficient labels from the target domain available.

\begin{table}
\centering
\caption{Comparison of domain-adaptive mixing strategies (SD: additional source domain data, S: data selection, TDM: target DepthMix, CDM: Cross-Domain DepthMix, MG: Matching Geometry Sampling, MTL: SDE multi-task learning). mIoU in \%, standard deviation over 3 seeds.}
\label{tab:ssda_advanced}
\setlength{\tabcolsep}{2.5pt}
\begin{tabular}{cccccccc}
\hline
SD  & S   & TDM  & CDM & MG  & MTL & 100 Trg. Labels & 500 Trg. Labels \\
\hline\hline
\cm &     &     &     &     &     & 53.83 \spm{1.09} & 60.99 \spm{1.04} \\
\cm &     & \cm &     &     &     & 60.05 \spm{1.91} & 66.19 \spm{0.80} \\
\cm &     &     & \cm &     &     & 60.65 \spm{1.88} & 65.34 \spm{0.08} \\
\cm &     & \cm & \cm &     &     & 61.35 \spm{1.39} & 66.98 \spm{0.88} \\
\cm &     & \cm & \cm & \cm &     & 63.00 \spm{2.09} & 67.14 \spm{0.42} \\
\hline
\cm & \cm & \cm & \cm & \cm & \cm & 66.01 \spm{0.32} & 69.88 \spm{0.39} \\
\hline
\end{tabular}
\vspace*{\floatsep}
\centering
\caption{Comparison of the class-wise IoU in \% of domain-adaptive mixing strategies for 100 labeled target samples (see \autoref{tab:ssda_advanced} for abbreviations). The color visualizes the IoU difference with respect to SD.}
\label{fig:ssda_advanced_classwise}
\includegraphics[width=1.0\linewidth]{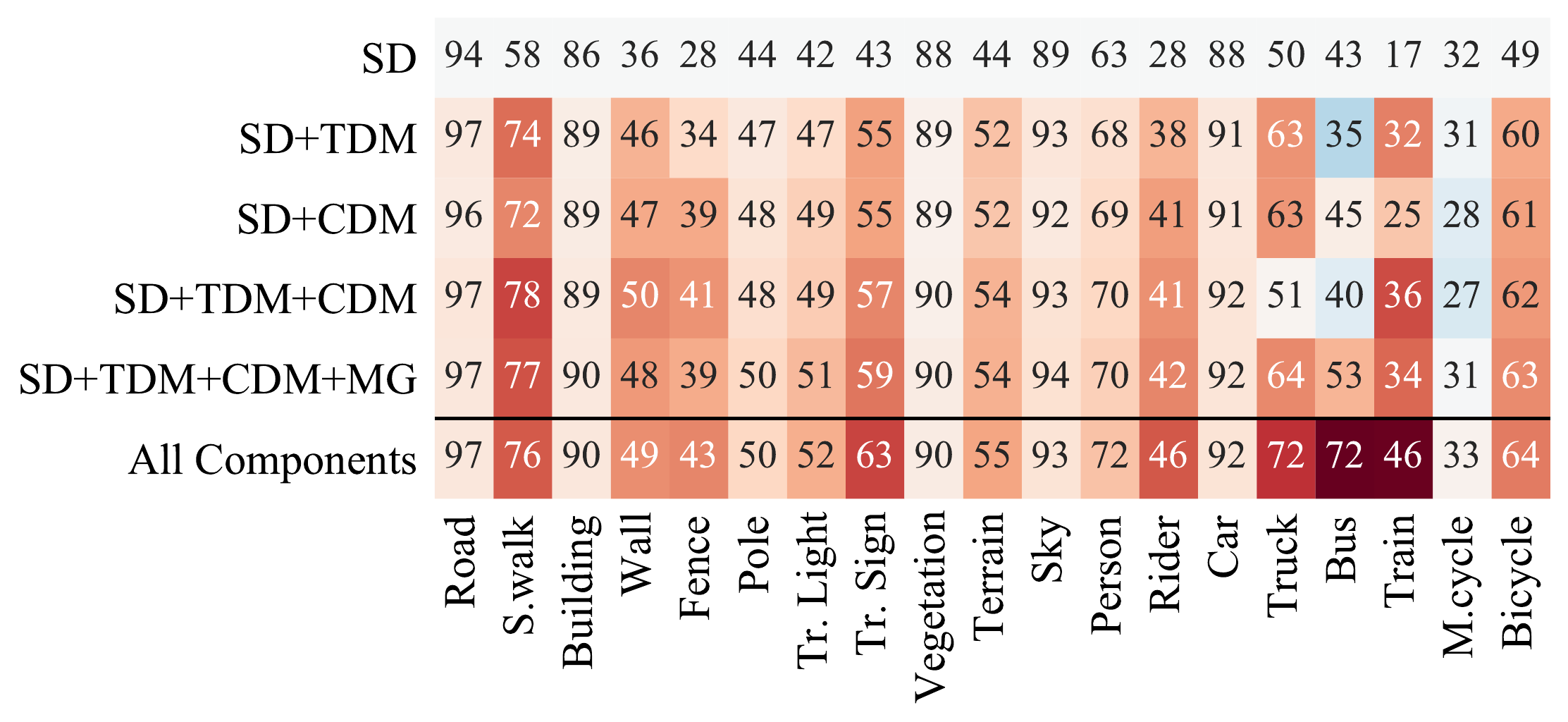}
\end{table}

Based on this observation, we conclude that it might be useful to combine CDM and TDM to align labeled source and target samples as well as labeled target and unlabeled target samples. \autoref{tab:ssda_advanced} shows that CDM+TDM indeed improves the performance over only CDM and only TDM by +0.70 (+0.79) for 100 (500) labeled target samples due to an improved performance for the classes sidewalk, wall, fence, traffic sign, terrain, and train.

To further improve the Cross-Domain DepthMix, we apply the proposed Matching Geometry Sampling to overcome the geometric domain gap of source and target domain and to better align the geometric distribution of the mixed samples to the geometric target distribution as discussed in \autoref{sec:methods_mg}. \autoref{tab:ssda_advanced} shows that it improves the mIoU by +1.65 (+0.16) percentage points for 100 (500) labeled target samples. 
The geometry and view alignment is probably more important for fewer labeled target samples because it is more difficult to bridge the geometric domain gap.
For 100 labeled samples, the improvement mainly originates from difficult vehicles such as truck, bus, and motorcycle (see \autoref{fig:ssda_advanced_classwise}).

When combining the domain adaptive strategies (combined CDM+TDM and Matching Geometry Sampling) with the previous contributions from SSL, the SSDA performance can be further improved by +3.01 (+2.74) percentage points for 100 (500) labeled target samples (see \autoref{tab:ssda_advanced}). Overall, our contributions sum up to +17.26 (+8.22) percentage points improvement over the baseline using only target supervision and +12.18 (+8.89) percentage points improvement over the baseline with target and source supervision. Especially, the performance of truck, bus, and train is increased by more than 50\% as shown in \autoref{fig:ssda_advanced_classwise}.

\subsection{Comparison with State-of-the-Art SSDA Methods}
\label{sec:ssda_sota}
    
\begin{table*}
\begin{minipage}{\textwidth}
\centering
\caption{Comparison with other SSDA methods for GTA $\rightarrow$ Cityscapes. The mIoU in \% on the Cityscapes validation set is shown for a different number of labeled target samples. Mean and standard deviation are aggregated over 3 random seeds. Additionally, the relative performance (Rel.) in~\% with respect to the fully-supervised baseline is shown. The best results are shown in bold font and the second-best results are underlined.}
\label{tab:ssda_comp_sota_gta}
\setlength{\tabcolsep}{2.6pt}
\begin{tabular}{|l|ll|ll|ll|ll|}
\hline
\# Labeled (Target)                             &  \multicolumn{2}{c|}{100}    & \multicolumn{2}{c|}{200}  & \multicolumn{2}{c|}{500} & \multicolumn{2}{c|}{2975}  \\
\hline
                                        & mIoU             & Rel.   & mIoU             & Rel.    & mIoU             & Rel.     & mIoU            & Rel. \\
\hline
Baseline \citep{wang2020alleviating}     & 43.6             &        & 47.1             &         & 53.6             &          & 65.9             & Ref.     \\
ASS \citep{wang2020alleviating}          & 54.2             & 82.3   & 56.0             & 85.0    & 60.2             & 91.4     & 69.1             & 104.9   \\
\hline
Baseline \citep{alonso2021semi}          & --           &        & --           &         & --           &          & 66.4             & Ref.     \\
\citet{alonso2021semi}      & 59.9         & 90.2   & 62.0         & 93.4    & 64.2         & 96.7     & --               &    \\
\hline
Baseline \citep{chen2021semi}            & 41.9             &         & 47.7             &        & 55.5             &          & 65.3             & Ref.     \\
\citet{chen2021semi}         & 61.2             & 93.7    & 60.5             & 92.6   & 64.3             & 98.5     & 69.8             & \textbf{106.9}    \\
\hline
Baseline                                & 48.75 \spm{1.52} &        & 54.04 \spm{0.64} &         & 61.66 \spm{0.90} &          & 67.77 \spm{0.13} & Ref.     \\
DACS \citep{tranheden2021dacs}\footnote{Results of the reimplementation in our experiment setting extending DACS from UDA to SSDA.} & 61.04 \spm{0.64} & 90.1   & 63.14 \spm{1.00} & 93.2    & 64.89 \spm{0.45} & 95.8     & 66.51 \spm{0.18} & 98.1    \\
Ours w/o Data Selection                 & \underline{64.14} \spm{1.96} & \underline{94.6}   & \underline{66.13} \spm{0.20} & \underline{97.6}    & 68.16 \spm{0.40} & \underline{100.6}    & \textbf{71.71} \spm{0.44} & \underline{105.8}     \\
Ours                                    & \textbf{66.01} \spm{0.32} & \textbf{97.4}   & \textbf{67.73} \spm{0.43} & \textbf{99.9}    & \textbf{69.88} \spm{0.39} & \textbf{103.1}    & --               &          \\
\hline
\end{tabular}
\end{minipage}
\end{table*}

\begin{table*}
\begin{minipage}{\textwidth}
\centering
\caption{Comparison with other SSDA methods for Synthia $\rightarrow$ Cityscapes. The mIoU in \% of 13 classes on the Cityscapes validation set is shown for a different number of labeled target samples. Mean and standard deviation are aggregated over 3 random seeds. Additionally, the relative performance (Rel.) in \% with respect to the fully-supervised baseline is shown. The best results are shown in bold font and the second-best results are underlined.}
\label{tab:ssda_comp_sota_synthia}
\setlength{\tabcolsep}{2.6pt}
\begin{tabular}{|l|ll|ll|ll|ll|}
\hline
\# Labeled (Target)                             &  \multicolumn{2}{c|}{100}    & \multicolumn{2}{c|}{200}  & \multicolumn{2}{c|}{500} & \multicolumn{2}{c|}{2975}  \\
\hline
                                                & mIoU                  & Rel.      & mIoU                  & Rel.      & mIoU                  & Rel.      & mIoU              & Rel.      \\
\hline
Baseline \citep{wang2020alleviating}            & 57.6			        &           & 60.8                  &           & 66.5		            &	        & 73.8              & Ref.      \\
ASS \citep{wang2020alleviating}                 & 62.1		            & 84.1	    & 64.8	                & 87.8	    & 69.8		            & 94.6	    & \s{77.1}          & 104.      \\
\hline
Baseline \citep{chen2021semi}                   & 53			        &           & 58.9	                &           & 61		            &           & 72.2	            & Ref.      \\
\citet{chen2021semi}                            & 68.4		            & \s{94.7}  & 69.8	                & 96.7	    & 71.7		            & 99.3	    & \f{77.2}          & \f{106.9} \\
\hline
Baseline                                        & 58.00 \spm{1.96}      &           & 63.26	\spm{0.91}  	&           & 67.74	\spm{0.48}	    &           & 73.34	\spm{0.21}  & Ref.      \\
DACS \citep{tranheden2021dacs}\footnote{Results of the reimplementation in our experiment setting extending DACS from UDA to SSDA.} & 64.88 \spm{0.30}      & 88.5	    & 67.72	\spm{1.19}	    & 92.3	    & 71.32	\spm{0.38}	    & 97.2	    & 74.43	\spm{0.41}	& 101.5     \\
Ours w/o Data Selection                         & \s{68.89} \spm{1.94}  & 93.9	    & \s{71.95}	\spm{0.49}  & \s{98.1}  & \s{74.06}	\spm{0.30}	& \s{101.0}	& 77.04	\spm{0.31}	& \s{105.0} \\
Ours                                            & \f{72.35} \spm{0.23}  & \f{98.7}	& \f{73.54}	\spm{0.67}  & \f{100.3} & \f{75.36}	\spm{0.26}	& \f{102.8}	& --				&           \\
\hline
\end{tabular}
\end{minipage}
\end{table*}

Finally, we compare our framework with other state-of-the-art SSDA methods on the benchmarks Synthia $\rightarrow$ Cityscapes (\autoref{tab:ssda_comp_sota_gta}) and GTA $\rightarrow$ Cityscapes (\autoref{tab:ssda_comp_sota_synthia}). For each method, its baseline performance is provided because the methods differ in their architecture and labeled subset. For better comparability between the architectures, we show the relative performance in \% with respect to the fully-supervised baseline. As the previous SSDA methods did not publish their implementation, labeled subset, or variance over the subset selection, we adapted the UDA state-of-the-art methods DACS \citep{tranheden2021dacs} to our framework for a fair comparison with a competitive method.

Considering the mIoU and the relative performance with respect to the fully-supervised baseline, our method noticeably outperforms the competitors for 100, 200, and 500 labeled target samples on both benchmarks.
Only in the fully-supervised case, \citet{chen2021semi} achieves slightly better results.
Moreover, it can be seen that even if we remove the data selection for annotation from our method, the previous statements still hold. 

\begin{figure*}
    \centering
    \includegraphics[width=\linewidth]{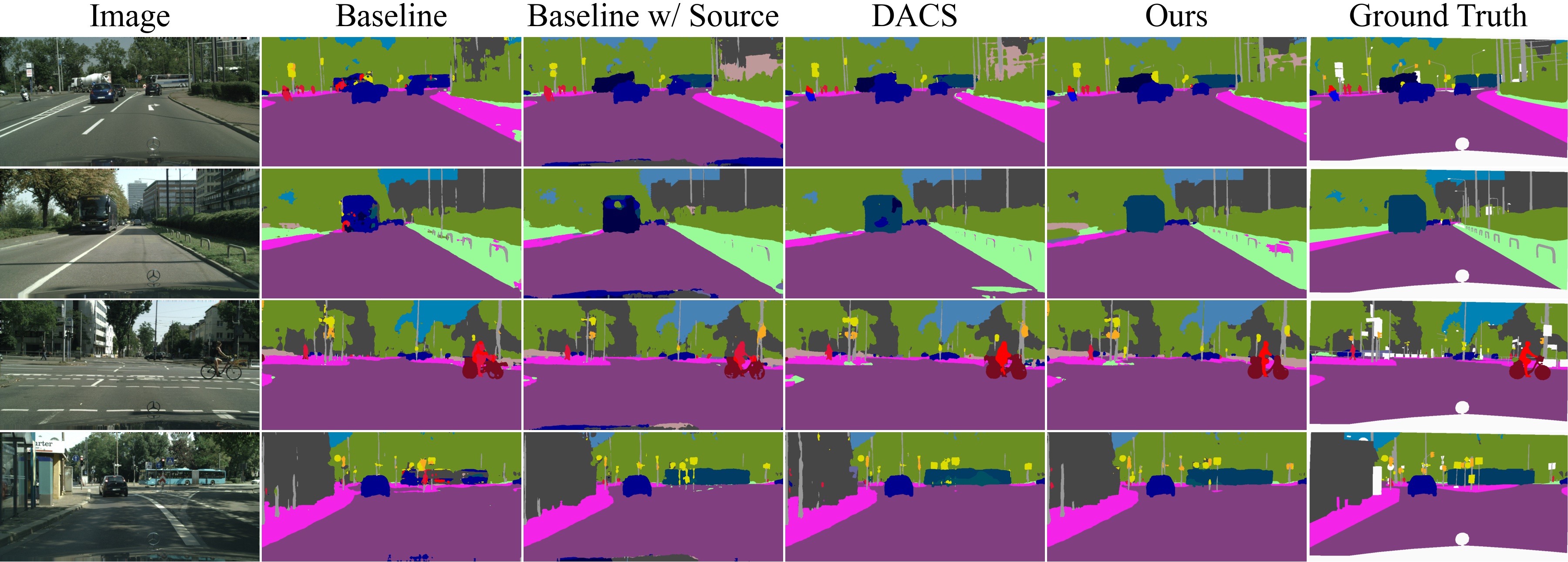}
    \caption{Example semantic segmentations from GTA5 $\rightarrow$ Cityscapes of our method for 100 labeled target samples in comparison with DACS \citep{tranheden2021dacs} adapted to SSDA and the baseline with/without source supervision.}
    \label{fig:examples_ssda}
\end{figure*}

We would like to highlight that our method achieves 97.4\% (GTA $\rightarrow$ Cityscapes) and 98.7\% (Synthia $\rightarrow$ Cityscapes) of the fully-supervised baseline performance with only about 1/30 (100) of the target labels. With about 1/15 of the target labels, it even reaches the fully-supervised baseline performance. The improved performance for 100 labeled target samples can also be observed in \autoref{fig:examples_ssda}, where our method better distinguishes difficult classes such as truck, bus, and train and produces more detailed segmentation contours for classes such as pole, traffic sign, and rider.

\section{Conclusions}
\label{sec:discussion}

In this work, we have studied how self-supervised depth estimation (SDE) can be utilized to improve semantic segmentation in the single-domain semi-supervised and the domain-adaptive semi-supervised setting.

We introduce four effective strategies capable of leveraging the knowledge learned from SDE. 
First, we present an automatic data selection for annotation algorithm based on SDE, which does not require human-in-the-loop annotations and, therefore, increases flexibility, efficiency, and scalability. By combining diversity sampling based on features from self-supervised depth estimation and uncertainty sampling based on the depth student error, our method significantly outperforms random data selection and even entropy-based active learning, which requires a human in the loop. We show that without knowledge of the class labels, our data selection for annotation prefers samples, which contain difficult/rare classes (e.g. rider, truck, bus, and train). This results in a significantly higher semantic segmentation performance of these classes.

Second, we demonstrate that the proposed DepthMix strategy outperforms related mixing strategies by avoiding an inconsistent geometry of the generated images. We show that DepthMix effectively improves the performance for classes with a low baseline performance such as wall, fence, traffic light, rider, truck, bus, and train. We assume that DepthMix improves generalization by presenting labeled and pseudo-labeled instances with different backgrounds and occlusions.

Third, we show that the feature representation from self-supervised depth estimation can be transferred to semantic segmentation, by means of SDE pretraining and multi-task learning of semantic segmentation and SDE. This is particularly effective for difficult classes surrounded by depth discontinuities such as wall, fence, pole, traffic, light, traffic sign, rider, truck, and motorcycle. By using an ImageNet feature distance loss during the SDE pretraining, we mitigate forgetting useful semantic features from ImageNet pretraining and avoid the resulting performance drop for semantically similar classes such as truck, bus, train, and motorcycle.

And fourth, we show the effectiveness of combined Cross-Domain and Target-Domain DepthMix as well as Matching Geometry Sampling in a semi-supervised domain adaptation setting. The former effectively aligns source and target data as well as labeled target and unlabeled data to generate high-quality pseudo-labels for unlabeled target data. The latter samples source images with a similar scene geometry and camera pose with respect to target images to produce more realistic Cross-Domain DepthMix images.

A combination of the first three contributions in a single-domain semi-supervised framework can achieve even higher performance gains than the single components as the approaches address different aspects of the learning process. By using these SDE-based contributions, our approach results in state-of-the-art performance for semi-supervised semantic segmentation. Our method achieves 92\% of the fully-supervised baseline performance with only 1/30 of the available labels and even slightly outperforms it with only 1/8 of the labels.

A combination of all four contributions in a semi\hyp{}supervised domain adaptation framework improves the performance even further and outperforms previous state-of-the-art semi-supervised domain adaptation methods. On GTA $\rightarrow$ Cityscapes, our method achieves even 97\% of the fully-supervised baseline performance with only 1/30 of the target labels. This roughly corresponds to only 150 working hours for data annotation for the target domain instead of 4460 working hours.

All in all, our findings suggest that SDE can be a valuable source of self-supervision for semantic segmentation, improving the semantic segmentation performance and reducing the number of necessary annotations.

\ifarxiv
\begin{acknowledgements}
This work is supported by Toyota Motor Europe via the research project TRACE-Zurich.
\end{acknowledgements}
\fi

\bibliographystyle{spbasic}      
\bibliography{references}   

\begin{thebibliography}{111}
\providecommand{\natexlab}[1]{#1}
\providecommand{\url}[1]{{#1}}
\providecommand{\urlprefix}{URL }
\expandafter\ifx\csname urlstyle\endcsname\relax
  \providecommand{\doi}[1]{DOI~\discretionary{}{}{}#1}\else
  \providecommand{\doi}{DOI~\discretionary{}{}{}\begingroup
  \urlstyle{rm}\Url}\fi
\providecommand{\eprint}[2][]{\url{#2}}

\bibitem[{Alonso et~al.(2021)Alonso, Sabater, Ferstl, Montesano, and
  Murillo}]{alonso2021semi}
Alonso I, Sabater A, Ferstl D, Montesano L, Murillo AC (2021) Semi-supervised
  semantic segmentation with pixel-level contrastive learning from a class-wise
  memory bank. arXiv preprint arXiv:210413415

\bibitem[{Araslanov and Roth(2021)}]{araslanov2021self}
Araslanov N, Roth S (2021) Self-supervised augmentation consistency for
  adapting semantic segmentation. In: IEEE Conf. Comput. Vis. Pattern Recog.,
  pp 15384--15394

\bibitem[{Berthelot et~al.(2019)Berthelot, Carlini, Goodfellow, Papernot,
  Oliver, and Raffel}]{berthelot2019mixmatch}
Berthelot D, Carlini N, Goodfellow I, Papernot N, Oliver A, Raffel CA (2019)
  Mixmatch: A holistic approach to semi-supervised learning. In: Adv. Neural
  Inform. Process. Syst., pp 5049--5059

\bibitem[{Brostow et~al.(2009)Brostow, Fauqueur, and
  Cipolla}]{brostow2009semantic}
Brostow GJ, Fauqueur J, Cipolla R (2009) Semantic object classes in video: A
  high-definition ground truth database. Pattern Recognition Letters
  30(2):88--97

\bibitem[{Casser et~al.(2019)Casser, Pirk, Mahjourian, and
  Angelova}]{casser2019depth}
Casser V, Pirk S, Mahjourian R, Angelova A (2019) Depth prediction without the
  sensors: Leveraging structure for unsupervised learning from monocular
  videos. In: AAAI Conf. Artif. Intell., pp 8001--8008

\bibitem[{Chapelle et~al.(2009)Chapelle, Scholkopf, and
  Zien}]{chapelle2009semi}
Chapelle O, Scholkopf B, Zien A (2009) Semi-supervised learning (chapelle, o.
  et al., eds.; 2006)[book reviews]. IEEE Trans on Neural Networks
  20(3):542--542

\bibitem[{Chen et~al.(2017)Chen, Papandreou, Kokkinos, Murphy, and
  Yuille}]{chen2017deeplab}
Chen LC, Papandreou G, Kokkinos I, Murphy K, Yuille AL (2017) Deeplab: Semantic
  image segmentation with deep convolutional nets, atrous convolution, and
  fully connected crfs. IEEE Trans Pattern Anal Mach Intell 40(4):834--848

\bibitem[{Chen et~al.(2019{\natexlab{a}})Chen, Liu, Liu, and
  Wang}]{chen2019towards}
Chen PY, Liu AH, Liu YC, Wang YCF (2019{\natexlab{a}}) Towards scene
  understanding: Unsupervised monocular depth estimation with semantic-aware
  representation. In: IEEE Conf. Comput. Vis. Pattern Recog., pp 2624--2632

\bibitem[{Chen et~al.(2021{\natexlab{a}})Chen, Jia, He, Shi, and
  Liu}]{chen2021semi}
Chen S, Jia X, He J, Shi Y, Liu J (2021{\natexlab{a}}) Semi-supervised domain
  adaptation based on dual-level domain mixing for semantic segmentation. In:
  IEEE Conf. Comput. Vis. Pattern Recog., pp 11018--11027

\bibitem[{Chen et~al.(2021{\natexlab{b}})Chen, Yuan, Zeng, and
  Wang}]{chen2021semia}
Chen X, Yuan Y, Zeng G, Wang J (2021{\natexlab{b}}) Semi-supervised semantic
  segmentation with cross pseudo supervision. In: IEEE Conf. Comput. Vis.
  Pattern Recog., pp 2613--2622

\bibitem[{Chen et~al.(2019{\natexlab{b}})Chen, Li, Chen, and
  Gool}]{chen2019learning}
Chen Y, Li W, Chen X, Gool LV (2019{\natexlab{b}}) Learning semantic
  segmentation from synthetic data: A geometrically guided input-output
  adaptation approach. In: IEEE Conf. Comput. Vis. Pattern Recog., pp
  1841--1850

\bibitem[{Chen et~al.(2019{\natexlab{c}})Chen, Schmid, and
  Sminchisescu}]{chen2019self}
Chen Y, Schmid C, Sminchisescu C (2019{\natexlab{c}}) Self-supervised learning
  with geometric constraints in monocular video: Connecting flow, depth, and
  camera. In: Int. Conf. Comput. Vis., pp 7063--7072

\bibitem[{Cordts et~al.(2016)Cordts, Omran, Ramos, Rehfeld, Enzweiler,
  Benenson, Franke, Roth, and Schiele}]{cordts2016cityscapes}
Cordts M, Omran M, Ramos S, Rehfeld T, Enzweiler M, Benenson R, Franke U, Roth
  S, Schiele B (2016) The cityscapes dataset for semantic urban scene
  understanding. In: IEEE Conf. Comput. Vis. Pattern Recog., pp 3213--3223

\bibitem[{Dai and Van~Gool(2018)}]{dai2018dark}
Dai D, Van~Gool L (2018) Dark model adaptation: Semantic image segmentation
  from daytime to nighttime. In: IEEE Int. Conf. on Intell. Transport. Syst.,
  pp 3819--3824

\bibitem[{Dai et~al.(2020)Dai, Patil, Hecker, Dai, Van~Gool, and
  Schindler}]{dai2020self}
Dai Q, Patil V, Hecker S, Dai D, Van~Gool L, Schindler K (2020) Self-supervised
  object motion and depth estimation from video. In: IEEE Conf. Comput. Vis.
  Pattern Recog. Workshops, pp 1004--1005

\bibitem[{Doersch et~al.(2015)Doersch, Gupta, and
  Efros}]{doersch2015unsupervised}
Doersch C, Gupta A, Efros AA (2015) Unsupervised visual representation learning
  by context prediction. In: Int. Conf. Comput. Vis., pp 1422--1430

\bibitem[{Dvornik et~al.(2019)Dvornik, Mairal, and
  Schmid}]{dvornik2019importance}
Dvornik N, Mairal J, Schmid C (2019) On the importance of visual context for
  data augmentation in scene understanding. IEEE Trans Pattern Anal Mach Intell

\bibitem[{Feng et~al.(2020{\natexlab{a}})Feng, Zhou, Cheng, Tan, Shi, and
  Ma}]{feng2020semi}
Feng Z, Zhou Q, Cheng G, Tan X, Shi J, Ma L (2020{\natexlab{a}})
  Semi-supervised semantic segmentation via dynamic self-training and
  class-balanced curriculum. arXiv preprint arXiv:200408514

\bibitem[{Feng et~al.(2020{\natexlab{b}})Feng, Zhou, Gu, Tan, Cheng, Lu, Shi,
  and Ma}]{fenga2020dmt}
Feng Z, Zhou Q, Gu Q, Tan X, Cheng G, Lu X, Shi J, Ma L (2020{\natexlab{b}})
  Dmt: Dynamic mutual training for semi-supervised learning. arXiv preprint
  arXiv:200408514

\bibitem[{French et~al.(2020)French, Laine, Aila, Mackiewicz, and
  Finlayson}]{french2019consistency}
French G, Laine S, Aila T, Mackiewicz M, Finlayson G (2020) Semi-supervised
  semantic segmentation needs strong, varied perturbations. In: Brit. Mach.
  Vis. Conf.

\bibitem[{Gal and Ghahramani(2016)}]{gal2016dropout}
Gal Y, Ghahramani Z (2016) Dropout as a bayesian approximation: Representing
  model uncertainty in deep learning. In: Int. Conf. Mach. Learning, pp
  1050--1059

\bibitem[{Garg et~al.(2016)Garg, BG, Carneiro, and Reid}]{garg2016unsupervised}
Garg R, BG VK, Carneiro G, Reid I (2016) Unsupervised cnn for single view depth
  estimation: Geometry to the rescue. In: Eur. Conf. Comput. Vis., pp 740--756

\bibitem[{Gidaris et~al.(2018)Gidaris, Singh, and
  Komodakis}]{gidaris2018unsupervised}
Gidaris S, Singh P, Komodakis N (2018) Unsupervised representation learning by
  predicting image rotations. In: Int. Conf. Learn. Represent.

\bibitem[{Godard et~al.(2017)Godard, Mac~Aodha, and
  Brostow}]{godard2017unsupervised}
Godard C, Mac~Aodha O, Brostow GJ (2017) Unsupervised monocular depth
  estimation with left-right consistency. In: IEEE Conf. Comput. Vis. Pattern
  Recog., pp 270--279

\bibitem[{Godard et~al.(2019)Godard, Mac~Aodha, Firman, and
  Brostow}]{godard2019digging}
Godard C, Mac~Aodha O, Firman M, Brostow GJ (2019) Digging into self-supervised
  monocular depth estimation. In: Int. Conf. Comput. Vis., pp 3828--3838

\bibitem[{Gonzalez~Bello and Kim(2020)}]{gonzalezbello2020forget}
Gonzalez~Bello JL, Kim M (2020) Forget about the lidar: Self-supervised depth
  estimators with med probability volumes. In: Adv. Neural Inform. Process.
  Syst.

\bibitem[{Goodfellow et~al.(2014)Goodfellow, Pouget-Abadie, Mirza, Xu,
  Warde-Farley, Ozair, Courville, and Bengio}]{goodfellow2014generative}
Goodfellow I, Pouget-Abadie J, Mirza M, Xu B, Warde-Farley D, Ozair S,
  Courville A, Bengio Y (2014) Generative adversarial nets. In: Adv. Neural
  Inform. Process. Syst., pp 2672--2680

\bibitem[{Gordon et~al.(2019)Gordon, Li, Jonschkowski, and
  Angelova}]{gordon2019depth}
Gordon A, Li H, Jonschkowski R, Angelova A (2019) Depth from videos in the
  wild: Unsupervised monocular depth learning from unknown cameras. In: Int.
  Conf. Comput. Vis., pp 8977--8986

\bibitem[{Guizilini et~al.(2020{\natexlab{a}})Guizilini, Ambrus, Pillai,
  Raventos, and Gaidon}]{guizilini20203d}
Guizilini V, Ambrus R, Pillai S, Raventos A, Gaidon A (2020{\natexlab{a}}) 3d
  packing for self-supervised monocular depth estimation. In: IEEE Conf.
  Comput. Vis. Pattern Recog., pp 2485--2494

\bibitem[{Guizilini et~al.(2020{\natexlab{b}})Guizilini, Hou, Li, Ambrus, and
  Gaidon}]{guizilini2020semantically}
Guizilini V, Hou R, Li J, Ambrus R, Gaidon A (2020{\natexlab{b}})
  Semantically-guided representation learning for self-supervised monocular
  depth. In: Int. Conf. Learn. Represent.

\bibitem[{Guizilini et~al.(2021)Guizilini, Li, Ambrus, and
  Gaidon}]{guizilini2021geometric}
Guizilini V, Li J, Ambrus R, Gaidon A (2021) Geometric unsupervised domain
  adaptation for semantic segmentation. arXiv preprint arXiv:210316694

\bibitem[{Hadsell et~al.(2006)Hadsell, Chopra, and
  LeCun}]{hadsell2006dimensionality}
Hadsell R, Chopra S, LeCun Y (2006) Dimensionality reduction by learning an
  invariant mapping. In: IEEE Conf. Comput. Vis. Pattern Recog., pp 1735--1742

\bibitem[{He et~al.(2016)He, Zhang, Ren, and Sun}]{he2016deep}
He K, Zhang X, Ren S, Sun J (2016) Deep residual learning for image
  recognition. In: IEEE Conf. Comput. Vis. Pattern Recog., pp 770--778

\bibitem[{He et~al.(2020)He, Fan, Wu, Xie, and Girshick}]{he2020momentum}
He K, Fan H, Wu Y, Xie S, Girshick R (2020) Momentum contrast for unsupervised
  visual representation learning. In: IEEE Conf. Comput. Vis. Pattern Recog.,
  pp 9729--9738

\bibitem[{Hoffman et~al.(2016)Hoffman, Wang, Yu, and Darrell}]{hoffman2016fcns}
Hoffman J, Wang D, Yu F, Darrell T (2016) Fcns in the wild: Pixel-level
  adversarial and constraint-based adaptation. arXiv preprint arXiv:161202649

\bibitem[{Hoffman et~al.(2018)Hoffman, Tzeng, Park, Zhu, Isola, Saenko, Efros,
  and Darrell}]{hoffman2018cycada}
Hoffman J, Tzeng E, Park T, Zhu JY, Isola P, Saenko K, Efros A, Darrell T
  (2018) Cycada: Cycle-consistent adversarial domain adaptation. In: Int. Conf.
  Mach. Learning, pp 1989--1998

\bibitem[{Hoyer et~al.(2021)Hoyer, Dai, Chen, Köring, Saha, and
  Van~Gool}]{hoyer2021three}
Hoyer L, Dai D, Chen Y, Köring A, Saha S, Van~Gool L (2021) Three ways to
  improve semantic segmentation with self-supervised depth estimation. In: IEEE
  Conf. Comput. Vis. Pattern Recog.

\bibitem[{Hung et~al.(2018)Hung, Tsai, Liou, Lin, and
  Yang}]{hung2018adversarial}
Hung WC, Tsai YH, Liou YT, Lin YY, Yang MH (2018) Adversarial learning for
  semi-supervised semantic segmentation. In: Brit. Mach. Vis. Conf.

\bibitem[{Huo et~al.(2021)Huo, Xie, He, Yang, Zhou, Li, and Tian}]{huo2021atso}
Huo X, Xie L, He J, Yang Z, Zhou W, Li H, Tian Q (2021) Atso: Asynchronous
  teacher-student optimization for semi-supervised image segmentation. In: IEEE
  Conf. Comput. Vis. Pattern Recog., pp 1235--1244

\bibitem[{Hwa(2004)}]{hwa2004sample}
Hwa R (2004) Sample selection for statistical parsing. Computational
  linguistics 30(3):253--276

\bibitem[{Ioffe and Szegedy(2015)}]{ioffe2015batch}
Ioffe S, Szegedy C (2015) Batch normalization: Accelerating deep network
  training by reducing internal covariate shift. arXiv preprint arXiv:150203167

\bibitem[{Jiang et~al.(2018)Jiang, Larsson, Maire Greg~Shakhnarovich, and
  Learned-Miller}]{jiang2018self}
Jiang H, Larsson G, Maire Greg~Shakhnarovich M, Learned-Miller E (2018)
  Self-supervised relative depth learning for urban scene understanding. In:
  Eur. Conf. Comput. Vis., pp 19--35

\bibitem[{Jiang et~al.(2019)Jiang, Sun, Jampani, Lv, Learned-Miller, and
  Kautz}]{jiang2019sense}
Jiang H, Sun D, Jampani V, Lv Z, Learned-Miller E, Kautz J (2019) Sense: A
  shared encoder network for scene-flow estimation. In: Int. Conf. Comput.
  Vis., pp 3195--3204

\bibitem[{Jiao et~al.(2018)Jiao, Cao, Song, and Lau}]{jiao2018look}
Jiao J, Cao Y, Song Y, Lau R (2018) Look deeper into depth: Monocular depth
  estimation with semantic booster and attention-driven loss. In: Eur. Conf.
  Comput. Vis., pp 53--69

\bibitem[{Kalluri et~al.(2019)Kalluri, Varma, Chandraker, and
  Jawahar}]{kalluri2019universal}
Kalluri T, Varma G, Chandraker M, Jawahar C (2019) Universal semi-supervised
  semantic segmentation. In: Int. Conf. Comput. Vis., pp 5259--5270

\bibitem[{Kasarla et~al.(2019)Kasarla, Nagendar, Hegde, Balasubramanian, and
  Jawahar}]{kasarla2019region}
Kasarla T, Nagendar G, Hegde GM, Balasubramanian V, Jawahar C (2019)
  Region-based active learning for efficient labeling in semantic segmentation.
  In: IEEE Winter Conf. Appl. of Comput. Vis., pp 1109--1117

\bibitem[{Kim and Byun(2020)}]{kim2020learning}
Kim M, Byun H (2020) Learning texture invariant representation for domain
  adaptation of semantic segmentation. In: IEEE Conf. Comput. Vis. Pattern
  Recog., pp 12975--12984

\bibitem[{Klingner et~al.(2020{\natexlab{a}})Klingner, Bar, and
  Fingscheidt}]{klingner2020improved}
Klingner M, Bar A, Fingscheidt T (2020{\natexlab{a}}) Improved noise and attack
  robustness for semantic segmentation by using multi-task training with
  self-supervised depth estimation. In: IEEE Conf. Comput. Vis. Pattern Recog.
  Workshops, pp 320--321

\bibitem[{Klingner et~al.(2020{\natexlab{b}})Klingner, Term{\"o}hlen,
  Mikolajczyk, and Fingscheidt}]{klingner2020self}
Klingner M, Term{\"o}hlen JA, Mikolajczyk J, Fingscheidt T (2020{\natexlab{b}})
  Self-supervised monocular depth estimation: Solving the dynamic object
  problem by semantic guidance. In: Eur. Conf. Comput. Vis., pp 582--600

\bibitem[{Lai et~al.(2021)Lai, Tian, Jiang, Liu, Zhao, Wang, and
  Jia}]{lai2021semi}
Lai X, Tian Z, Jiang L, Liu S, Zhao H, Wang L, Jia J (2021) Semi-{Supervised}
  {Semantic} {Segmentation} {With} {Directional} {Context}-{Aware}
  {Consistency}. In: CVPR, pp 1205--1214

\bibitem[{LeCun et~al.(1998)LeCun, Bottou, Bengio, and
  Haffner}]{lecun1998gradient}
LeCun Y, Bottou L, Bengio Y, Haffner P (1998) Gradient-based learning applied
  to document recognition. Proceedings of the IEEE 86(11):2278--2324

\bibitem[{Lee(2013)}]{lee2013pseudo}
Lee DH (2013) Pseudo-label: The simple and efficient semi-supervised learning
  method for deep neural networks. In: Int. Conf. Mach. Learning

\bibitem[{Lee et~al.(2018)Lee, Ros, Li, and Gaidon}]{lee2018spigan}
Lee KH, Ros G, Li J, Gaidon A (2018) Spigan: Privileged adversarial learning
  from simulation. In: Int. Conf. Learn. Represent.

\bibitem[{Li et~al.(2018)Li, Wang, Dong, Yan, Liu, and Zha}]{li2018joint}
Li C, Wang X, Dong W, Yan J, Liu Q, Zha H (2018) Joint active learning with
  feature selection via cur matrix decomposition. IEEE Trans Pattern Anal Mach
  Intell 41(6):1382--1396

\bibitem[{Li et~al.(2020{\natexlab{a}})Li, Ma, Kang, Yuan, Zhang, and
  Wang}]{li2020deep}
Li C, Ma H, Kang Z, Yuan Y, Zhang XY, Wang G (2020{\natexlab{a}}) On deep
  unsupervised active learning. Int Joint Conf Artif Intell

\bibitem[{Li et~al.(2020{\natexlab{b}})Li, Kang, Liu, Wei, and
  Yang}]{li2020content}
Li G, Kang G, Liu W, Wei Y, Yang Y (2020{\natexlab{b}}) Content-consistent
  matching for domain adaptive semantic segmentation. In: Eur. Conf. Comput.
  Vis., pp 440--456

\bibitem[{Lian et~al.(2019)Lian, Lv, Duan, and Gong}]{lian2019constructing}
Lian Q, Lv F, Duan L, Gong B (2019) Constructing self-motivated pyramid
  curriculums for cross-domain semantic segmentation: A non-adversarial
  approach. In: Int. Conf. Comput. Vis., pp 6758--6767

\bibitem[{Long et~al.(2015)Long, Shelhamer, and Darrell}]{long2015fully}
Long J, Shelhamer E, Darrell T (2015) Fully convolutional networks for semantic
  segmentation. In: IEEE Conf. Comput. Vis. Pattern Recog., pp 3431--3440

\bibitem[{Mackowiak et~al.(2018)Mackowiak, Lenz, Ghori, Diego, Lange, and
  Rother}]{mackowiak2018cereals}
Mackowiak R, Lenz P, Ghori O, Diego F, Lange O, Rother C (2018)
  Cereals-cost-effective region-based active learning for semantic
  segmentation. In: Brit. Mach. Vis. Conf.

\bibitem[{McCallumzy and Nigamy(1998)}]{mccallumzy1998employing}
McCallumzy AK, Nigamy K (1998) Employing em and pool-based active learning for
  text classification. In: Int. Conf. Mach. Learning, pp 359--367

\bibitem[{Mendel et~al.(2020)Mendel, De~Souza, Rauber, Papa, and
  Palm}]{mendel2020semi}
Mendel R, De~Souza LA, Rauber D, Papa JP, Palm C (2020) Semi-supervised
  segmentation based on error-correcting supervision. In: Eur. Conf. Comput.
  Vis., pp 141--157

\bibitem[{Mittal et~al.(2019)Mittal, Tatarchenko, and Brox}]{mittal2019semi}
Mittal S, Tatarchenko M, Brox T (2019) Semi-supervised semantic segmentation
  with high-and low-level consistency. IEEE Trans Pattern Anal Mach Intell

\bibitem[{Nie et~al.(2013)Nie, Wang, Huang, and Ding}]{nie2013early}
Nie F, Wang H, Huang H, Ding C (2013) Early active learning via robust
  representation and structured sparsity. In: Int. Joint Conf. Artif. Intell.

\bibitem[{Novosel et~al.(2019)Novosel, Viswanath, and
  Arsenali}]{novoselboosting}
Novosel J, Viswanath P, Arsenali B (2019) Boosting semantic segmentation with
  multi-task self-supervised learning for autonomous driving applications. In:
  Int. Conf. Comput. Vis. Workshops

\bibitem[{Olsson et~al.(2021)Olsson, Tranheden, Pinto, and
  Svensson}]{olsson2020classmix}
Olsson V, Tranheden W, Pinto J, Svensson L (2021) Classmix: Segmentation-based
  data augmentation for semi-supervised learning. In: IEEE Winter Conf. on
  Applications of Comput. Vis., pp 1369--1378

\bibitem[{Ouali et~al.(2020)Ouali, Hudelot, and Tami}]{ouali2020semi}
Ouali Y, Hudelot C, Tami M (2020) Semi-supervised semantic segmentation with
  cross-consistency training. In: IEEE Conf. Comput. Vis. Pattern Recog., pp
  12674--12684

\bibitem[{Pilzer et~al.(2018)Pilzer, Xu, Puscas, Ricci, and
  Sebe}]{pilzer2018unsupervised}
Pilzer A, Xu D, Puscas M, Ricci E, Sebe N (2018) Unsupervised adversarial depth
  estimation using cycled generative networks. In: Int. Conf. on 3D Vision, pp
  587--595

\bibitem[{Pilzer et~al.(2019)Pilzer, Lathuiliere, Sebe, and
  Ricci}]{pilzer2019refine}
Pilzer A, Lathuiliere S, Sebe N, Ricci E (2019) Refine and distill: Exploiting
  cycle-inconsistency and knowledge distillation for unsupervised monocular
  depth estimation. In: IEEE Conf. Comput. Vis. Pattern Recog., pp 9768--9777

\bibitem[{Ramirez et~al.(2018)Ramirez, Poggi, Tosi, Mattoccia, and
  Di~Stefano}]{ramirez2018geometry}
Ramirez PZ, Poggi M, Tosi F, Mattoccia S, Di~Stefano L (2018) Geometry meets
  semantics for semi-supervised monocular depth estimation. In: Asian Conf.
  Comput. Vis., pp 298--313

\bibitem[{Ramirez et~al.(2019)Ramirez, Tonioni, Salti, and
  Stefano}]{ramirez2019learning}
Ramirez PZ, Tonioni A, Salti S, Stefano LD (2019) Learning across tasks and
  domains. In: Int. Conf. Comput. Vis., pp 8110--8119

\bibitem[{Ranjan et~al.(2019)Ranjan, Jampani, Balles, Kim, Sun, Wulff, and
  Black}]{ranjan2019competitive}
Ranjan A, Jampani V, Balles L, Kim K, Sun D, Wulff J, Black MJ (2019)
  Competitive collaboration: Joint unsupervised learning of depth, camera
  motion, optical flow and motion segmentation. In: IEEE Conf. Comput. Vis.
  Pattern Recog., pp 12240--12249

\bibitem[{Richter et~al.(2016)Richter, Vineet, Roth, and
  Koltun}]{richter2016playing}
Richter SR, Vineet V, Roth S, Koltun V (2016) Playing for data: Ground truth
  from computer games. In: Eur. Conf. Comput. Vis., pp 102--118

\bibitem[{Richter et~al.(2017)Richter, Hayder, and
  Koltun}]{richter2017playing_sequences}
Richter SR, Hayder Z, Koltun V (2017) Playing for benchmarks. In: Int. Conf.
  Comput. Vis., pp 2213--2222

\bibitem[{Ronneberger et~al.(2015)Ronneberger, Fischer, and
  Brox}]{ronneberger2015u}
Ronneberger O, Fischer P, Brox T (2015) U-net: Convolutional networks for
  biomedical image segmentation. In: Int. Conf. Medical Image Computing and
  Computer-assisted Intervention, pp 234--241

\bibitem[{Ros et~al.(2016)Ros, Sellart, Materzynska, Vazquez, and
  Lopez}]{germansellart2016large}
Ros G, Sellart L, Materzynska J, Vazquez D, Lopez AM (2016) A large collection
  of synthetic images for semantic segmentation of urban scenes. In: IEEE Conf.
  Comput. Vis. Pattern Recog., pp 3234--3243

\bibitem[{Sakaridis et~al.(2018)Sakaridis, Dai, and
  Van~Gool}]{sakaridis2018semantic}
Sakaridis C, Dai D, Van~Gool L (2018) Semantic foggy scene understanding with
  synthetic data. Int J Comput Vis 126(9):973--992

\bibitem[{Sakaridis et~al.(2021)Sakaridis, Dai, and
  Van~Gool}]{sakaridis2021acdc}
Sakaridis C, Dai D, Van~Gool L (2021) {ACDC}: The adverse conditions dataset
  with correspondences for semantic driving scene understanding. In: Int. Conf.
  Comput. Vis.

\bibitem[{Sener and Savarese(2018)}]{sener2017active}
Sener O, Savarese S (2018) Active learning for convolutional neural networks: A
  core-set approach. In: Int. Conf. Learn. Represent.

\bibitem[{Settles(2009)}]{settles2009active}
Settles B (2009) Active learning literature survey. Tech. rep., University of
  Wisconsin-Madison Department of Computer Sciences

\bibitem[{Settles and Craven(2008)}]{settles2008analysis}
Settles B, Craven M (2008) An analysis of active learning strategies for
  sequence labeling tasks. In: Conf. Empirical Methods Natural Language
  Processing, pp 1070--1079

\bibitem[{Seung et~al.(1992)Seung, Opper, and Sompolinsky}]{seung1992query}
Seung HS, Opper M, Sompolinsky H (1992) Query by committee. In: Annual Workshop
  Computational Learning Theory, pp 287--294

\bibitem[{Shu et~al.(2020)Shu, Yu, Duan, and Yang}]{shu2020feature}
Shu C, Yu K, Duan Z, Yang K (2020) Feature-metric loss for self-supervised
  learning of depth and egomotion. In: Eur. Conf. Comput. Vis., pp 572--588

\bibitem[{Siddiqui et~al.(2020)Siddiqui, Valentin, and
  Nie{\ss}ner}]{siddiqui2020viewal}
Siddiqui Y, Valentin J, Nie{\ss}ner M (2020) Viewal: Active learning with
  viewpoint entropy for semantic segmentation. In: IEEE Conf. Comput. Vis.
  Pattern Recog., pp 9433--9443

\bibitem[{Sinha et~al.(2019)Sinha, Ebrahimi, and
  Darrell}]{sinha2019variational}
Sinha S, Ebrahimi S, Darrell T (2019) Variational adversarial active learning.
  In: Int. Conf. Comput. Vis., pp 5972--5981

\bibitem[{Sohn et~al.(2020)Sohn, Berthelot, Carlini, Zhang, Zhang, Raffel,
  Cubuk, Kurakin, and Li}]{sohn2020fixmatch}
Sohn K, Berthelot D, Carlini N, Zhang Z, Zhang H, Raffel CA, Cubuk ED, Kurakin
  A, Li CL (2020) Fixmatch: Simplifying semi-supervised learning with
  consistency and confidence. In: Adv. Neural Inform. Process. Syst.

\bibitem[{Souly et~al.(2017)Souly, Spampinato, and Shah}]{souly2017semi}
Souly N, Spampinato C, Shah M (2017) Semi supervised semantic segmentation
  using generative adversarial network. In: Int. Conf. Comput. Vis., pp
  5688--5696

\bibitem[{Tarvainen and Valpola(2017)}]{tarvainen2017mean}
Tarvainen A, Valpola H (2017) Mean teachers are better role models:
  Weight-averaged consistency targets improve semi-supervised deep learning
  results. In: Adv. Neural Inform. Process. Syst., pp 1195--1204

\bibitem[{Tranheden et~al.(2021)Tranheden, Olsson, Pinto, and
  Svensson}]{tranheden2021dacs}
Tranheden W, Olsson V, Pinto J, Svensson L (2021) Dacs: Domain adaptation via
  cross-domain mixed sampling. In: IEEE Winter Conf. on Applications of Comput.
  Vis., pp 1379--1389

\bibitem[{Tsai et~al.(2018)Tsai, Hung, Schulter, Sohn, Yang, and
  Chandraker}]{tsai2018learning}
Tsai YH, Hung WC, Schulter S, Sohn K, Yang MH, Chandraker M (2018) Learning to
  adapt structured output space for semantic segmentation. In: IEEE Conf.
  Comput. Vis. Pattern Recog., pp 7472--7481

\bibitem[{Vandenhende et~al.(2021)Vandenhende, Georgoulis, Van~Gansbeke,
  Proesmans, Dai, and Van~Gool}]{vandenhende2020revisiting}
Vandenhende S, Georgoulis S, Van~Gansbeke W, Proesmans M, Dai D, Van~Gool L
  (2021) Multi-task learning for dense prediction tasks: A survey. IEEE Trans
  Pattern Anal Mach Intell

\bibitem[{Verma et~al.(2019)Verma, Lamb, Kannala, Bengio, and
  Lopez-Paz}]{verma2019interpolation}
Verma V, Lamb A, Kannala J, Bengio Y, Lopez-Paz D (2019) Interpolation
  consistency training for semi-supervised learning. In: Int. Joint Conf.
  Artif. Intell., pp 3635--3641

\bibitem[{Vu et~al.(2019{\natexlab{a}})Vu, Jain, Bucher, Cord, and
  P{\'e}rez}]{vu2019advent}
Vu TH, Jain H, Bucher M, Cord M, P{\'e}rez P (2019{\natexlab{a}}) Advent:
  Adversarial entropy minimization for domain adaptation in semantic
  segmentation. In: IEEE Conf. Comput. Vis. Pattern Recog., pp 2517--2526

\bibitem[{Vu et~al.(2019{\natexlab{b}})Vu, Jain, Bucher, Cord, and
  P{\'e}rez}]{vu2019dada}
Vu TH, Jain H, Bucher M, Cord M, P{\'e}rez P (2019{\natexlab{b}}) Dada:
  Depth-aware domain adaptation in semantic segmentation. In: Int. Conf.
  Comput. Vis., pp 7364--7373

\bibitem[{Wang et~al.(2021)Wang, Dai, Hoyer, Fink, and
  Van~Gool}]{wang2021domain}
Wang Q, Dai D, Hoyer L, Fink O, Van~Gool L (2021) Domain adaptive semantic
  segmentation with self-supervised depth estimation. In: Int. Conf. Comput.
  Vis.

\bibitem[{Wang et~al.(2019)Wang, Pizer, and Frahm}]{wang2019recurrent}
Wang R, Pizer SM, Frahm JM (2019) Recurrent neural network for (un-) supervised
  learning of monocular video visual odometry and depth. In: IEEE Conf. Comput.
  Vis. Pattern Recog., pp 5555--5564

\bibitem[{Wang et~al.(2020)Wang, Wei, Feris, Xiong, Hwu, Huang, and
  Shi}]{wang2020alleviating}
Wang Z, Wei Y, Feris R, Xiong J, Hwu WM, Huang TS, Shi H (2020) Alleviating
  semantic-level shift: A semi-supervised domain adaptation method for semantic
  segmentation. In: IEEE Conf. Comput. Vis. Pattern Recog. Workshops, pp
  936--937

\bibitem[{Wei et~al.(2018)Wei, Xiao, Shi, Jie, Feng, and
  Huang}]{wei2018revisiting}
Wei Y, Xiao H, Shi H, Jie Z, Feng J, Huang TS (2018) Revisiting dilated
  convolution: A simple approach for weakly-and semi-supervised semantic
  segmentation. In: IEEE Conf. Comput. Vis. Pattern Recog., pp 7268--7277

\bibitem[{Xie et~al.(2020)Xie, Feng, Chen, Sun, Ma, and Song}]{xie2020deal}
Xie S, Feng Z, Chen Y, Sun S, Ma C, Song M (2020) Deal: Difficulty-aware active
  learning for semantic segmentation. In: Asian Conf. Comput. Vis.

\bibitem[{Xu et~al.(2018)Xu, Ouyang, Wang, and Sebe}]{xu2018pad}
Xu D, Ouyang W, Wang X, Sebe N (2018) Pad-net: Multi-tasks guided
  prediction-and-distillation network for simultaneous depth estimation and
  scene parsing. In: IEEE Conf. Comput. Vis. Pattern Recog., pp 675--684

\bibitem[{Yang et~al.(2018)Yang, Zhao, Shi, Deng, and Jia}]{yang2018segstereo}
Yang G, Zhao H, Shi J, Deng Z, Jia J (2018) Segstereo: Exploiting semantic
  information for disparity estimation. In: Eur. Conf. Comput. Vis., pp
  636--651

\bibitem[{Yang et~al.(2017)Yang, Zhang, Chen, Zhang, and
  Chen}]{yang2017suggestive}
Yang L, Zhang Y, Chen J, Zhang S, Chen DZ (2017) Suggestive annotation: A deep
  active learning framework for biomedical image segmentation. In: Int. Conf.
  Medical Image Computing and Computer-assisted Intervention, pp 399--407

\bibitem[{Yang and Soatto(2020)}]{yang2020fda}
Yang Y, Soatto S (2020) Fda: Fourier domain adaptation for semantic
  segmentation. In: IEEE Conf. Comput. Vis. Pattern Recog., pp 4085--4095

\bibitem[{Yin and Shi(2018)}]{yin2018geonet}
Yin Z, Shi J (2018) Geonet: Unsupervised learning of dense depth, optical flow
  and camera pose. In: IEEE Conf. Comput. Vis. Pattern Recog., pp 1983--1992

\bibitem[{Yu et~al.(2006)Yu, Bi, and Tresp}]{yu2006active}
Yu K, Bi J, Tresp V (2006) Active learning via transductive experimental
  design. In: Int. Conf. Mach. Learning, pp 1081--1088

\bibitem[{Yun et~al.(2019)Yun, Han, Oh, Chun, Choe, and Yoo}]{yun2019cutmix}
Yun S, Han D, Oh SJ, Chun S, Choe J, Yoo Y (2019) Cutmix: Regularization
  strategy to train strong classifiers with localizable features. In: Int.
  Conf. Comput. Vis., pp 6023--6032

\bibitem[{Zhang et~al.(2021)Zhang, Zhang, Zhang, Chen, Wang, and
  Wen}]{zhang2021prototypical}
Zhang P, Zhang B, Zhang T, Chen D, Wang Y, Wen F (2021) Prototypical pseudo
  label denoising and target structure learning for domain adaptive semantic
  segmentation. In: Proceedings of the IEEE/CVF Conference on Computer Vision
  and Pattern Recognition, pp 12414--12424

\bibitem[{Zhang et~al.(2019)Zhang, David, Foroosh, and
  Gong}]{zhang2019curriculum}
Zhang Y, David P, Foroosh H, Gong B (2019) A curriculum domain adaptation
  approach to the semantic segmentation of urban scenes. IEEE Trans Pattern
  Anal Mach Intell 42(8):1823--1841

\bibitem[{Zheng et~al.(2019)Zheng, Yang, Chen, Han, Zhang, Liang, Zhao, Wang,
  and Chen}]{zheng2019biomedical}
Zheng H, Yang L, Chen J, Han J, Zhang Y, Liang P, Zhao Z, Wang C, Chen DZ
  (2019) Biomedical image segmentation via representative annotation. In: AAAI
  Conf. Artif. Intell., pp 5901--5908

\bibitem[{Zheng and Yang(2021)}]{zheng2021rectifying}
Zheng Z, Yang Y (2021) Rectifying pseudo label learning via uncertainty
  estimation for domain adaptive semantic segmentation. Int J Comput Vis
  129(4):1106--1120

\bibitem[{Zhou et~al.(2017)Zhou, Brown, Snavely, and
  Lowe}]{zhou2017unsupervised}
Zhou T, Brown M, Snavely N, Lowe DG (2017) Unsupervised learning of depth and
  ego-motion from video. In: IEEE Conf. Comput. Vis. Pattern Recog., pp
  1851--1858

\bibitem[{Zou et~al.(2018)Zou, Yu, Kumar, and Wang}]{zou2018unsupervised}
Zou Y, Yu Z, Kumar B, Wang J (2018) Unsupervised domain adaptation for semantic
  segmentation via class-balanced self-training. In: Eur. Conf. Comput. Vis.,
  pp 289--305

\end{thebibliography}

\end{document}